\documentclass[letterpaper, 10 pt, conference]{ieeeconf}  

\IEEEoverridecommandlockouts                              
\usepackage{graphicx}
\usepackage{siunitx}
\usepackage[utf8]{inputenc}
\usepackage[T1]{fontenc}
\usepackage{amsfonts}
\newcommand{\fig}[1]{Fig.~\ref{#1}}

\newcommand{\eq}[1]{(\ref{#1})}

\usepackage{epsfig} 
\usepackage{amsmath} 
\usepackage{subcaption}
\usepackage{caption}
\usepackage{color}
\usepackage{verbatim}
\usepackage{wasysym}
\usepackage[table,xcdraw]{xcolor}
\usepackage{graphicx}
\usepackage{gensymb}
\usepackage{xcolor}

\newcommand{\norm}[1]{\| #1 \|}
\newcommand{\abs}[1]{\lvert #1 \rvert}
\newcommand{\RNum}[1]{\uppercase\expandafter{\romannumeral #1\relax}}
\makeatletter
\newlength\tmp@\newlength\t@mp
\newcommand{\comp}[3]
  {\mathop{ \settowidth\tmp@{$\displaystyle\mathop{#1}^{#3}_{#2}$}
  \hbox to \tmp@{\hss \settowidth\t@mp{$\displaystyle #1$}\setlength\t@mp{.45\t@mp}
  $\displaystyle\mathop{#1}^{\hspace\t@mp #3}_{\hspace{-\t@mp}#2}$
  \hss} }}
\makeatother
\title{\LARGE \bf
An Under-Actuated Whippletree Mechanism Gripper based on Multi-Objective Design Optimization with Auto-Tuned Weights
}

\author{Yusuke Tanaka$^{1}$, Yuki Shirai$^{1}$, Zachary Lacey$^{1}$, Xuan Lin$^{1}$, Jane Liu$^{1}$, and Dennis Hong$^{1}$
\thanks{$^{1}$Y. Tanaka, Y. Shirai, Z. Lacey, X. Lin, J. Liu and D. Hong are with the Robotics and Mechanisms Laboratory, Department of Mechanical and Aerospace Engineering, University of California, Los Angeles, CA 90095, USA.
       {\tt\small yusuketanaka@g.ucla.edu}, {\tt\small yukishirai4869@g.ucla.edu}, {\tt\small zlacey@g.ucla.edu}, {\tt\small maynight@ucla.edu}, {\tt\small liujane@g.ucla.edu}, {\tt\small dennishong@ucla.edu}
        }
}

\begin{document}
\maketitle
\thispagestyle{empty}
\pagestyle{empty}

\begin{abstract}
Current rigid linkage grippers are limited in flexibility, and gripper design optimality relies on expertise, experiments, or arbitrary parameters. 
Our proposed rigid gripper can accommodate irregular and off-center objects through a whippletree mechanism, improving adaptability. We present a whippletree-based rigid under-actuated gripper and its parametric design multi-objective optimization for a one-wall climbing task. Our proposed objective function considers kinematics and grasping forces simultaneously with a mathematical metric based on a model of an object environment.
Our multi-objective problem is formulated as a single kinematic objective function with auto-tuning force-based weight. Our results indicate that our proposed objective function determines optimal parameters and kinematic ranges for our under-actuated gripper in the task environment with sufficient grasping forces. 

\end{abstract}

\section{INTRODUCTION}

The use of versatile end-effectors has extended beyond factories, finding use in cooking, farming, and climbing applications.
However, designing general under-actuated grippers for organic and variable object shapes requires expert and experienced engineering knowledge to decide the workspace or degree of versatility. 
Design decisions can significantly affect a gripper design's optimality, resulting in grippers optimized for arbitrary kinematic ranges \cite{kinematic_optimization}, or maximum grasping forces \cite{underactuated_tendon}. 
In addition, a particular design's adaptability is often tested via a limited trial-and-error basis rather than through mathematical formulation. A mathematical approach can diminish the expertise required to design a gripper and ensure reproducible results from design optimization. Furthermore, the design's generality is not limited to the extent of conducted experiments. It is infeasible to try all possible shapes of objects, but those sizes can be expressed in a distribution function.

Parametric design optimizations have been done against several metrics that characterize gripper capabilities mathematically. Suitable metrics such as kinematics, force ratio, or force isotropy are selected based on tasks \cite{orient_opti}, or grasping objects and calculated analytically \cite{surgerry_opti}. Simulation-based objective function evaluations can include metrics not obtainable through analytical means, such as success rates and uncertainties \cite{sensitive_grasp}. 
However, physics simulations increase the computational expense of function evaluations, and analysis is limited to object geometries tested in the simulations. 

 Optimizing a gripper design for kinematics and grasping forces simultaneously requires a multi-objective function.
 Multi-objective problems (MOP) often conflict as improving one aspect of the objective may worsen others, and the relative relationships between different objectives need to be determined by educated guesses \cite{surgerry_opti}, statistical approaches \cite{multiobject_design}, or by finding the Pareto front through an evolutionary algorithm \cite{opti_evolutionary}.

 \begin{figure}[t!]
    \centering
    \includegraphics[width=0.49\textwidth, trim ={2cm 2.5cm 0.3cm 2cm},clip]{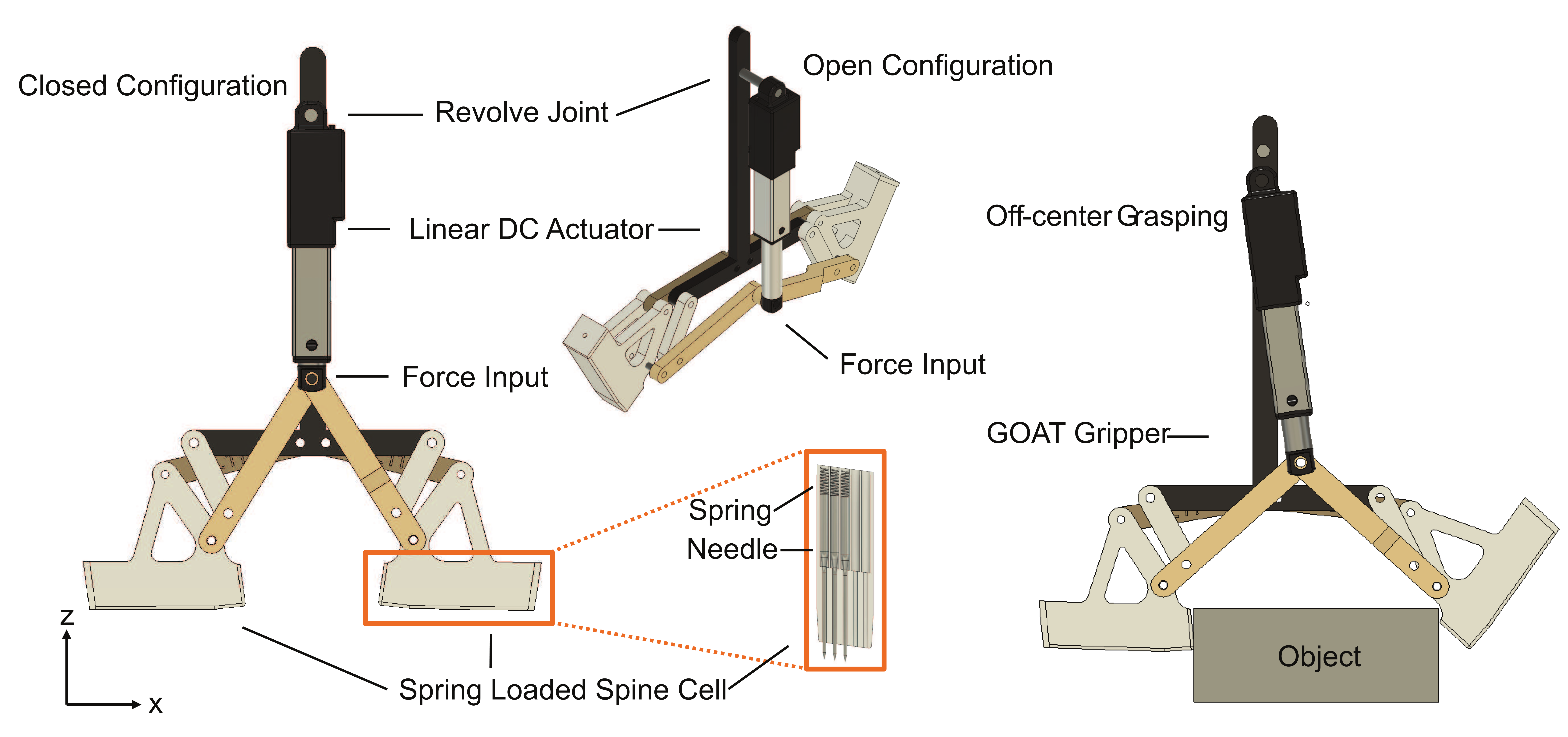}
     \caption{Our optimal whippletree-mechanism under-actuated adaptable rigid gripper. Spring loaded spines are employed at the tip to decrease the likelihood of slipping off the  bouldering holds. The topology of the 2D design is shown in \fig{fig:whipple_design} and \fig{fig:gripper_parameters}\label{fig:cad}}
\end{figure} 
 
 Rigid grippers have been widely adopted in industries and real-world applications due to their simplicity, low maintenance, and higher grasping forces \cite{industrial_gripper}. Despite these economic benefits, rigid mechanisms often lack the flexibility and versatility. Robotiq and parallel jaw grippers have to grasp an object in their center, whereas Barrethand can grasp objects off center by adding more degrees of freedom with a pulley mechanism \cite{underactuated_tendon}. More adaptability was achieved by tendon-driven grippers such as  Spinyhand \cite{spinyhand}, a soft Gecko elastomer actuator gripper \cite{gekko_soft}, and a soft and tendon-driven hybrid approach \cite{wire-driven_soft}.
 Adaptive rigid grippers have been proposed to meet several different criteria, such as traversability \cite{orient_opti} shifting between parallel and encompassing grip \cite{adaptable_gripper}. 
However, those linkage-based grippers lack robustness against the operating environment's uncertainties. 

 A gripper has to compensate offsets from an object to grasp due to various uncertainties such as vision pose estimations \cite{vision_opti}, and accommodate geometry deviations of the objects \cite{robust_opti}. A whipple (whiffle) tree mechanism can mechanically distribute load. It is commonly used in adaptable tendon-driven under-actuated grippers \cite{tendon_whiffle}, but can be applied to a rigid system \cite{whippletree_rigid}. This mechanically intelligent system can maintain gripper rigidity and simplicity while adding adaptability and robustness.

This paper presents an under-actuated whippletree-based rigid two-finger gripper and a multi-objective optimization. 
Our mechanically intelligent whippletree mechanism can provide the adaptability and robustness with rigid linkage mechanism and one actuator, which have only been achieved by soft or tendon-driven grippers.
The best design parameters are chosen through an multi-objective optimization on kinematics and grasping force using auto-tuning weights.
We demonstrate the capability of the proposed design and the objective function in the case of a one-wall climbing robot that needs to grasp bouldering holds.


%

The contributions are summarized as follows:
\begin{enumerate}
    \item We propose a rigid under-actuated gripper based on a whippletree mechanism, which can passively adapt and compensate for an off-center object.
    \item We formulate the environment-based objective function that considers both kinematic adaptability and grasping force. 
    \item We propose an auto-tuning weighting function to represent nonlinear relationships between kinematics and forces.
    \item We validate our proposed under-actuated gripper and objective function in hardware experiments.
\end{enumerate}

\section{PROBLEM FORMULATION\label{sec:pf}}
This section describes the design of our rigid two whippletree-based gripper kinematics, models of the environment, and a position-controlled limbed robot toe position accuracy.

\subsection{Design of the Rigid Whippletree-based Gripper \label{sec:design}}
\begin{figure}
 \centering
    \begin{subfigure}{0.2\textwidth}
    
\includegraphics[width=\textwidth,trim={0cm -1cm 0cm 0cm}, clip]{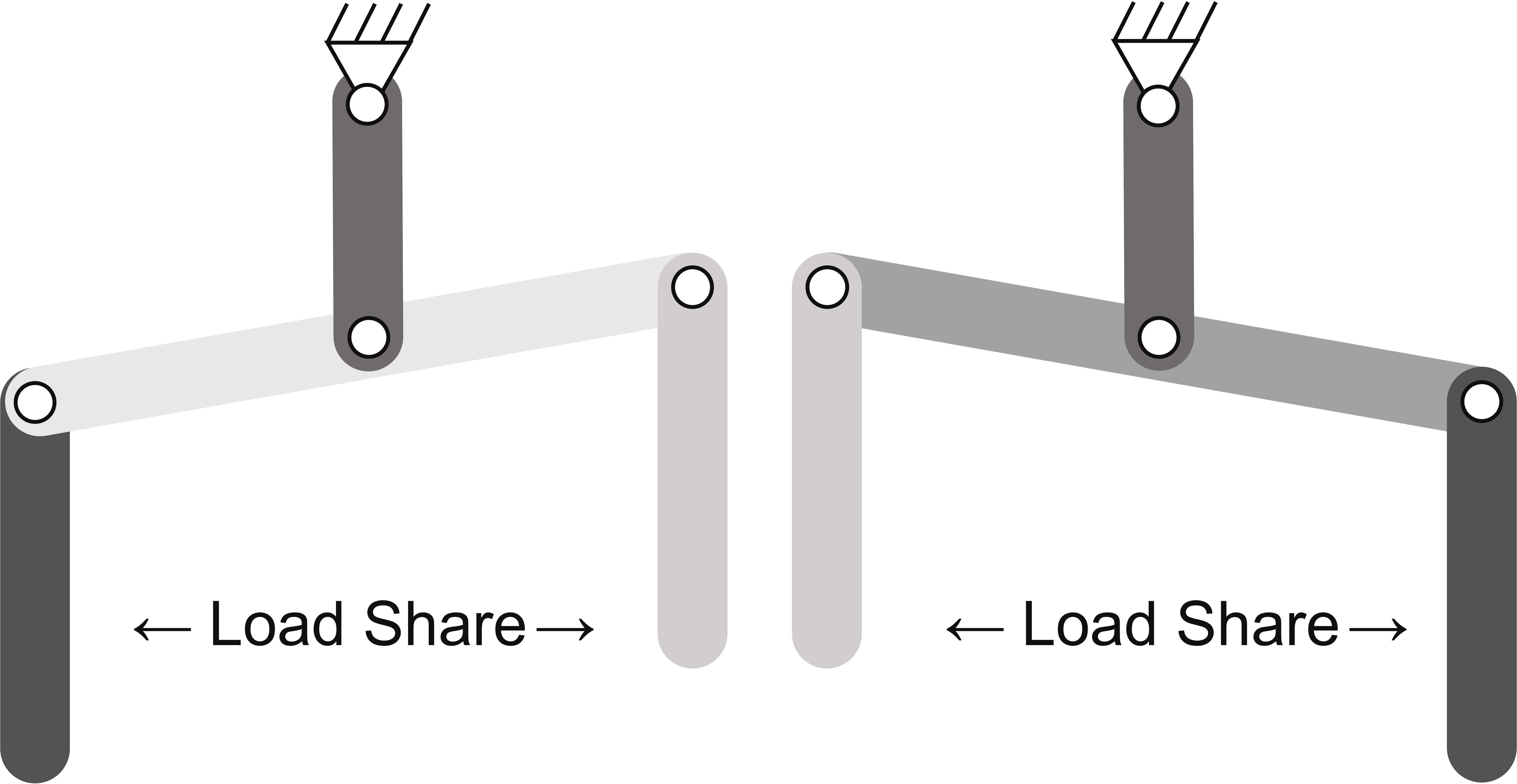}
\caption{The basis design of our GOAT. Two conventional whippletrees.}
    \end{subfigure}
     \begin{subfigure}{0.2\textwidth}
         \centering
    \includegraphics[width=\textwidth, clip]{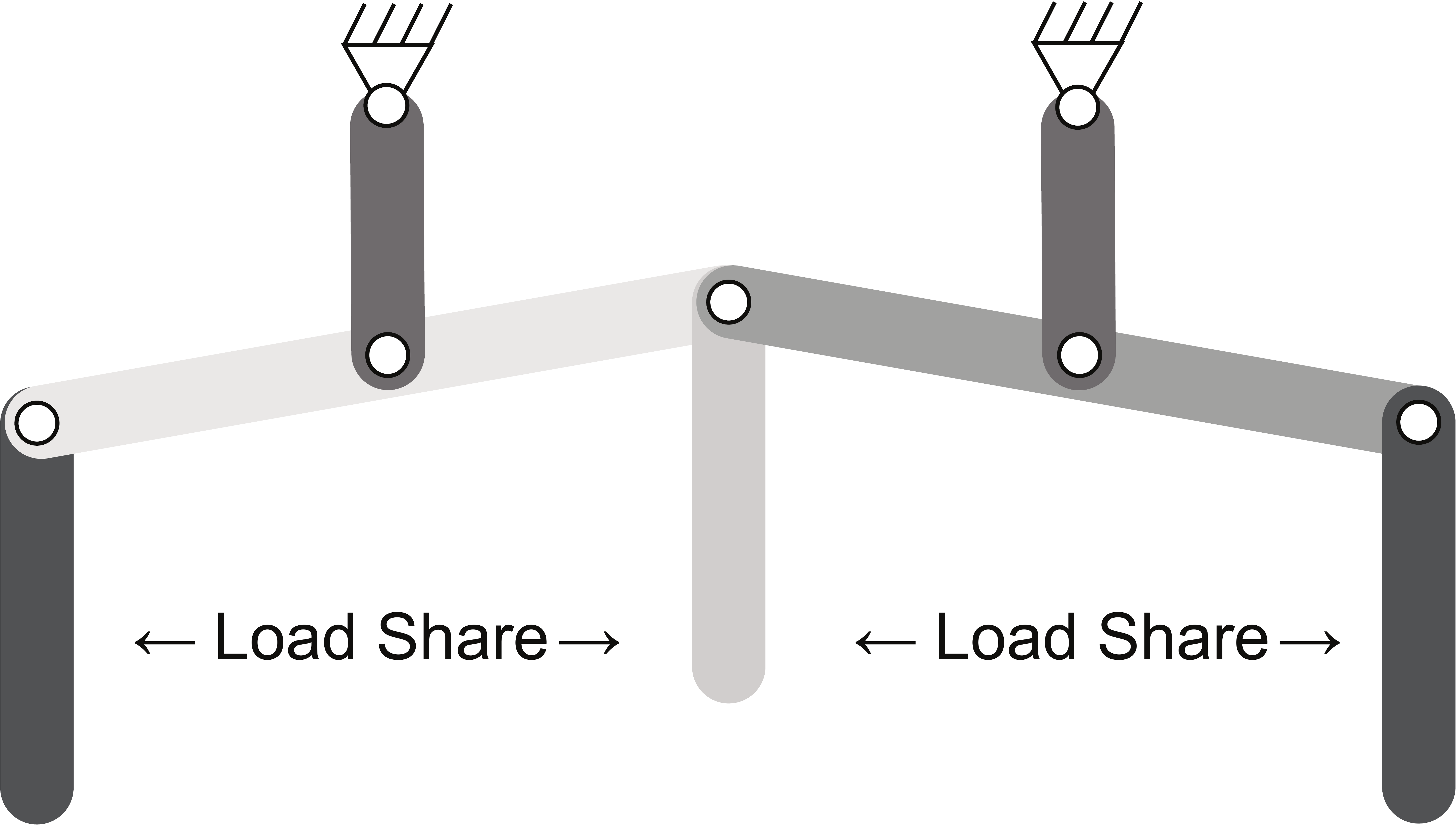}
    \caption{one load link combined from each whippletree mechanisms.}
     \end{subfigure}
      \begin{subfigure}{0.2\textwidth}
    \centering
\includegraphics[width=\textwidth, clip]{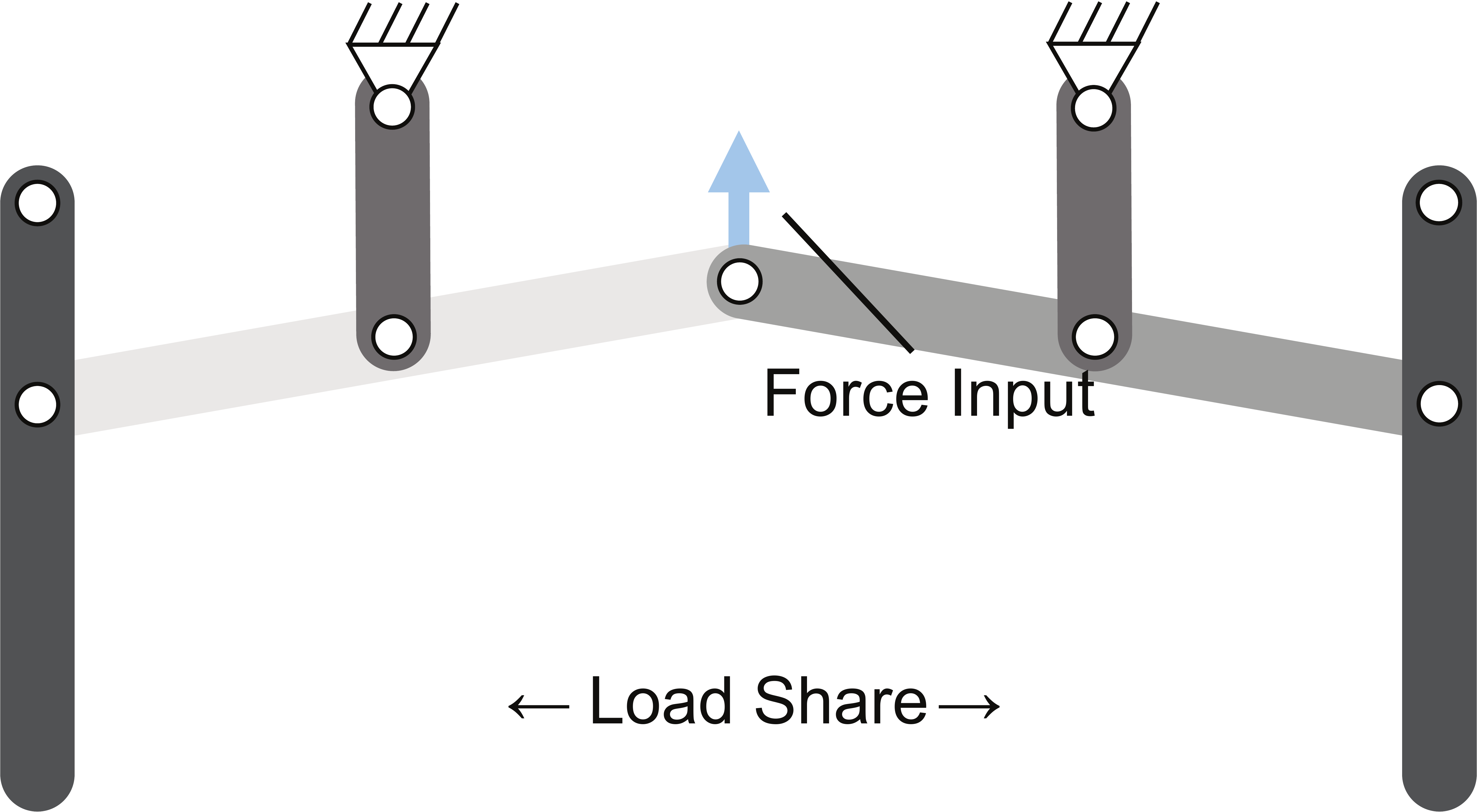}
\caption{The middle shared link is used as our input to the gripper, which force is distributed and passively balanced between the two links.}
    \end{subfigure}
     \begin{subfigure}{0.2\textwidth}
         \centering
    \includegraphics[width=\textwidth, clip]{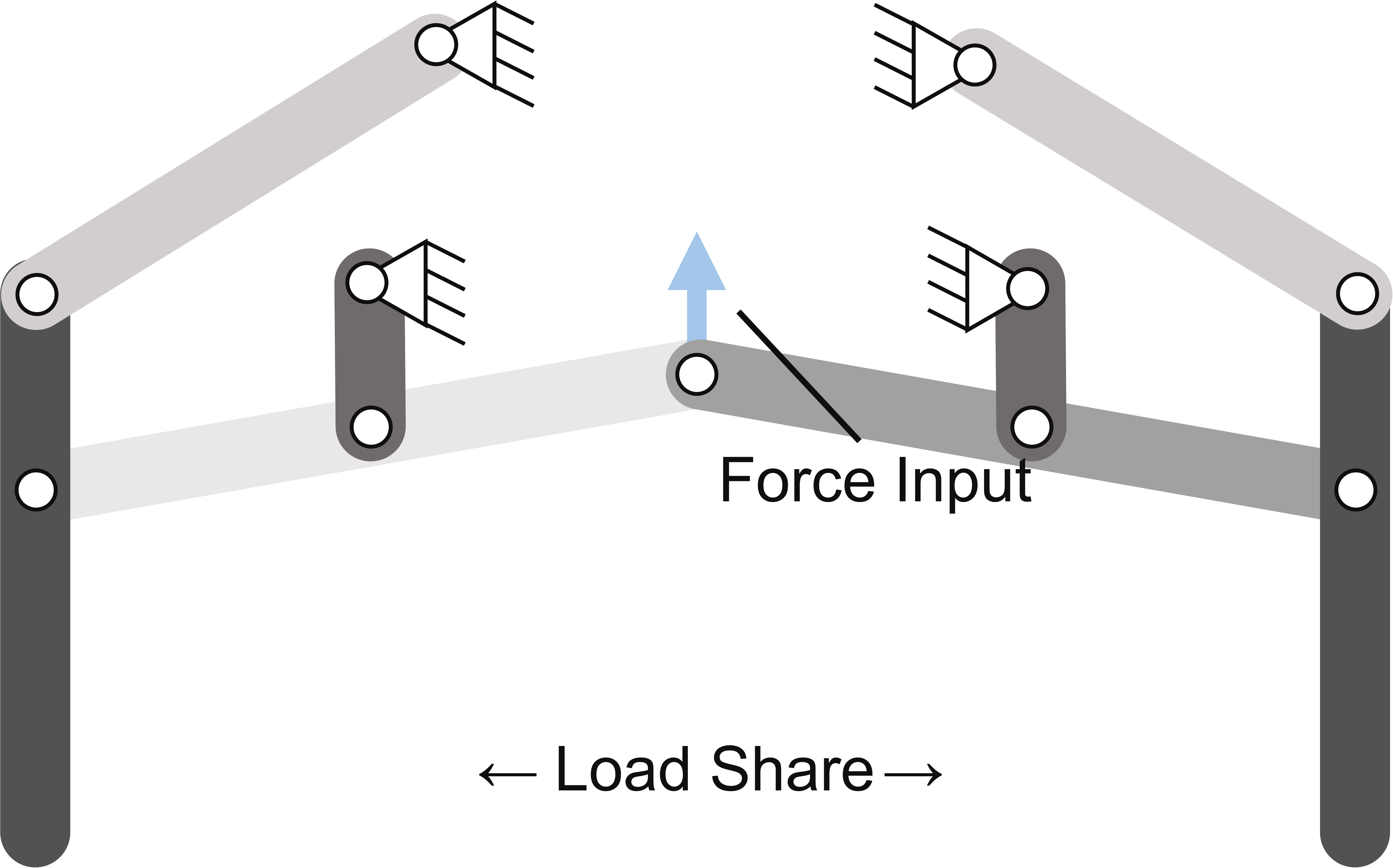}
    \caption{The load balanced two links are the gripper fingertips, which are constrained by adding two links.}
     \end{subfigure}
     \caption{Two whippletree-based gripper design evolution. Each individual member is represented by a straight line and revolve joints are denoted by circles. Two whippletrees share one link that is to input force on the load balancing mechanism. Two linkages are added to constraint the output links.
     }\label{fig:whipple_design}
\end{figure}

\begin{figure}
    \centering
    \includegraphics[width=0.35\textwidth,trim={1cm 0 1cm 0}, clip]{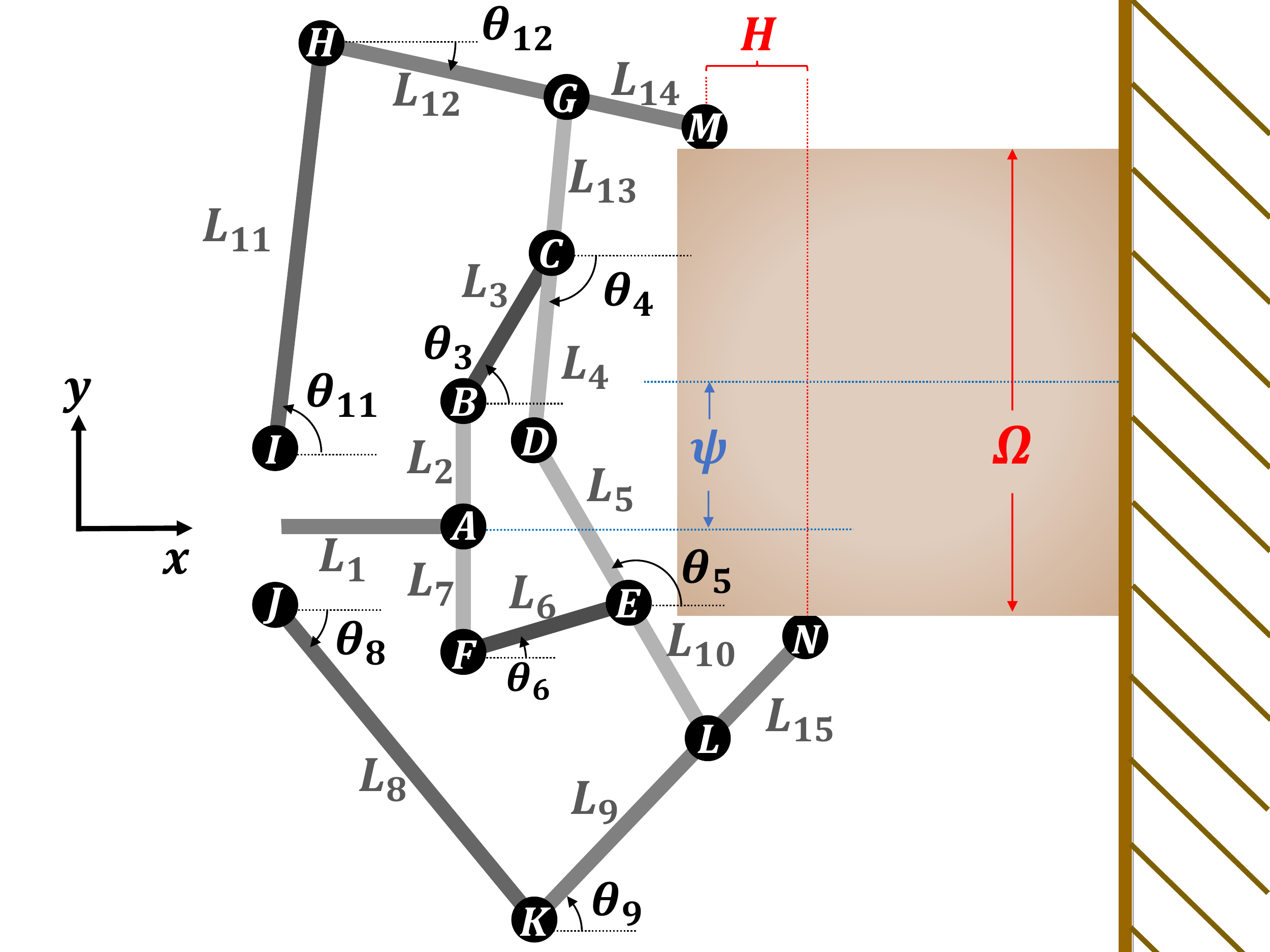}
    \caption{GOAT kinematic definitions. The point M and N are the fingertip positions. The differences in $x$ and $y$ represent the minimum hold height $H$ and the width graspable $\Omega$. The point D is a force input point. $L_1,L_2$ and $L_7$ are one body and grounded to a robot toe. The points I and J are fixed on $L_1$. $L_5$ and $L_{10}$, $L_4$ and $L_{13}$, $L_9$ and $L_{15}$, and $L_{12}$ and $L_{14}$ form single members respectively. These link lengths are non-optimal. }  
    \label{fig:gripper_parameters}
\end{figure}

Our proposed rigid under-actuated gripper, GOAT (Grasp Onto Any Terrain), CAD models, design and topology are shown in \fig{fig:cad}, \fig{fig:whipple_design}, and \fig{fig:gripper_parameters}, respectively. One whippletree has two output points: two links that are load-balanced. We combine one output link from each of the two whippletrees. One load applied at the shared joint is distributed between the two output links which are the gripper's fingertips. 
GOAT includes two five-bar linkages jointed at the point D which provides two constraints to the system. Hence from Maxwell's equations, GOAT has total two degrees of freedom. Then, the input point D can move in $x$ and $y$ Cartesian coordinate with no contact at the tips. One end of the linear actuator is fixed at the point D and the other is grounded at the revolve joint labeled in \fig{fig:cad}. Therefore, the linear actuator controls the point D in in spherical coordinate system.  If one finger touches an object surface the remaining finger will continue moving inward. This behavior allows it to passively adapt to an off-center object and evenly distribute one actuator force between two fingers. 
GOAT yields one exact kinematic solution given the point D position where its $x$-axis movement controls width between finger tips and its $y$-axis movement depends on the center of grasping. The under-actuated GOAT is stable when grasping a rigid object with no slip since the point D cannot purely move in $y$ direction where the liner actuator length is fix.  
 The finger tip force is calculated using static force equilibrium and we model the maximum pulling forces that GOAT can withstand through experiments in Section \ref{sec:GP}. One challenge here is that we have to carefully decide all link lengths to meet task requirements and ensure appropriate adaptability. These requirements are addressed by the multi-objective kinematic and force optimization described in Section \ref{sec:Opti}.



\subsection{Modeling of the Object Environment \label{sec:environment}}
\begin{figure}[t!]
 \centering
    \begin{subfigure}{0.2\textwidth}
\includegraphics[width=\textwidth, trim = {1.5cm 0 2.5cm 0}, clip]{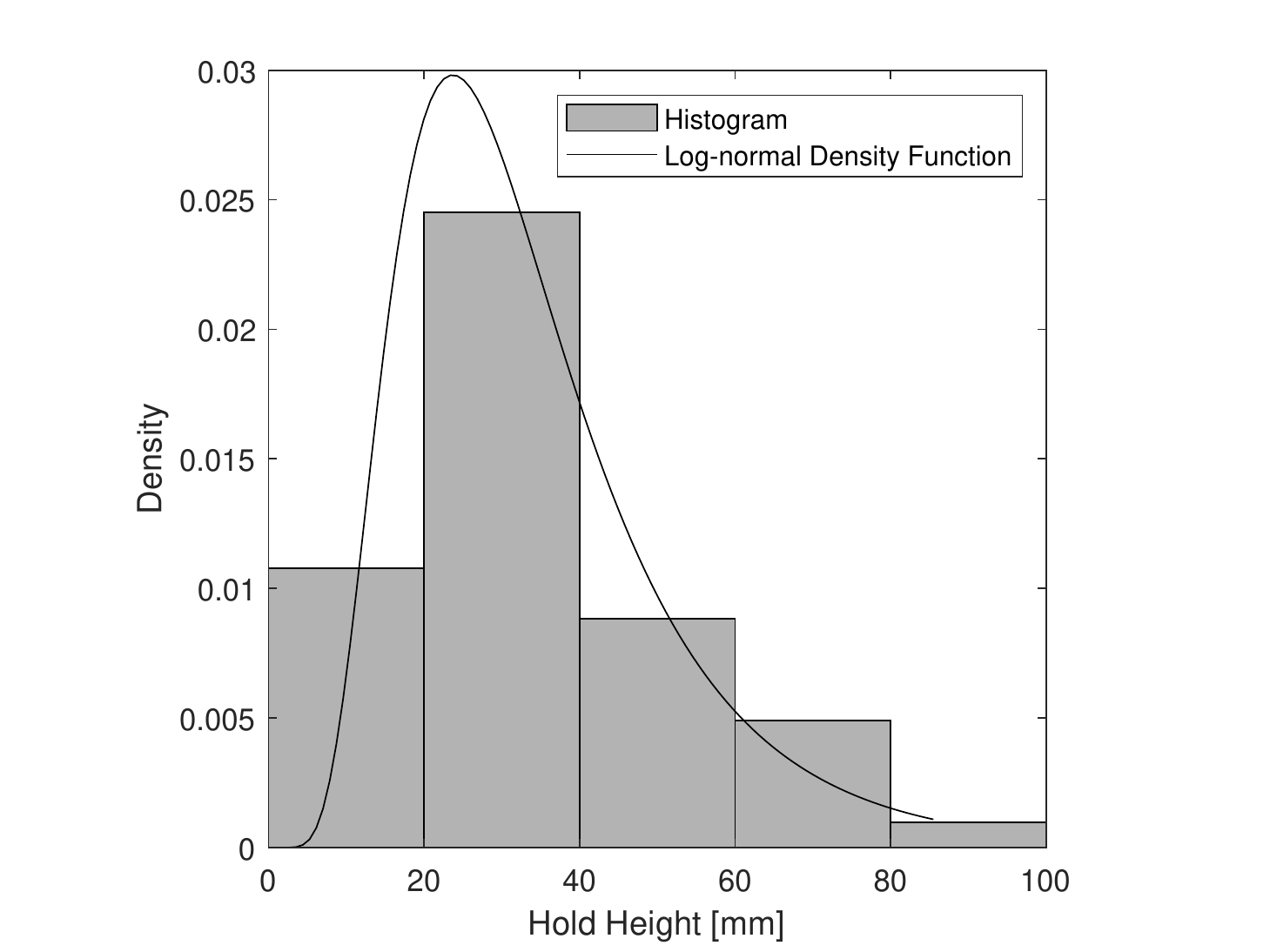}
\caption{Histogram and fitted log-normal distribution for hold height. \label{fig:histogram_height}}
    \end{subfigure}
     \begin{subfigure}{0.2\textwidth}
         \centering
    \includegraphics[width=\textwidth,trim = {1.5cm 0 2.5cm 0}, clip]{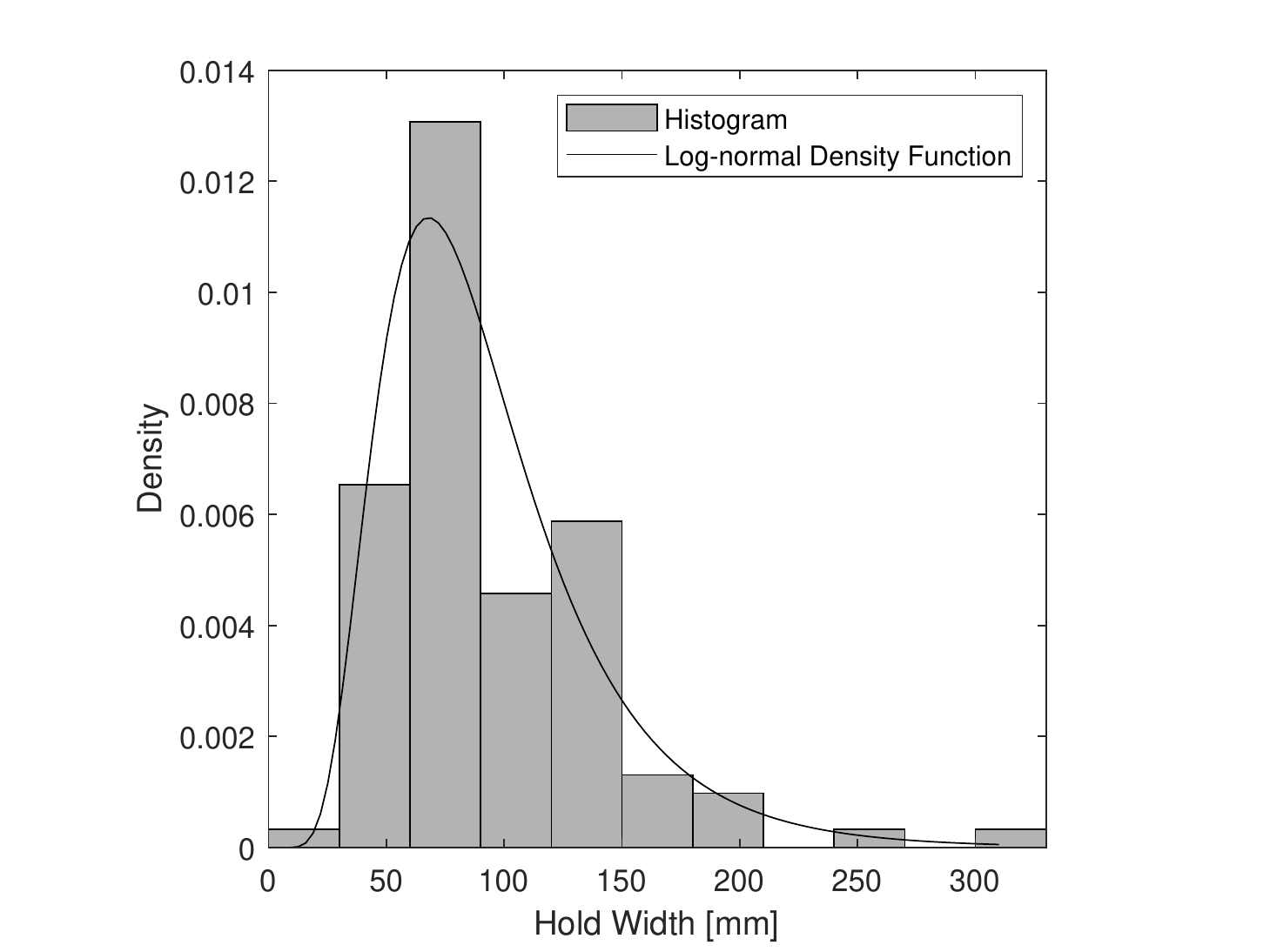}
    \caption{Histogram and fitted log-normal distribution for hold width. \label{fig:histogram_width}}
     \end{subfigure}
          \begin{subfigure}{0.5\textwidth}
         \centering
    \includegraphics[width=0.8\textwidth, clip]{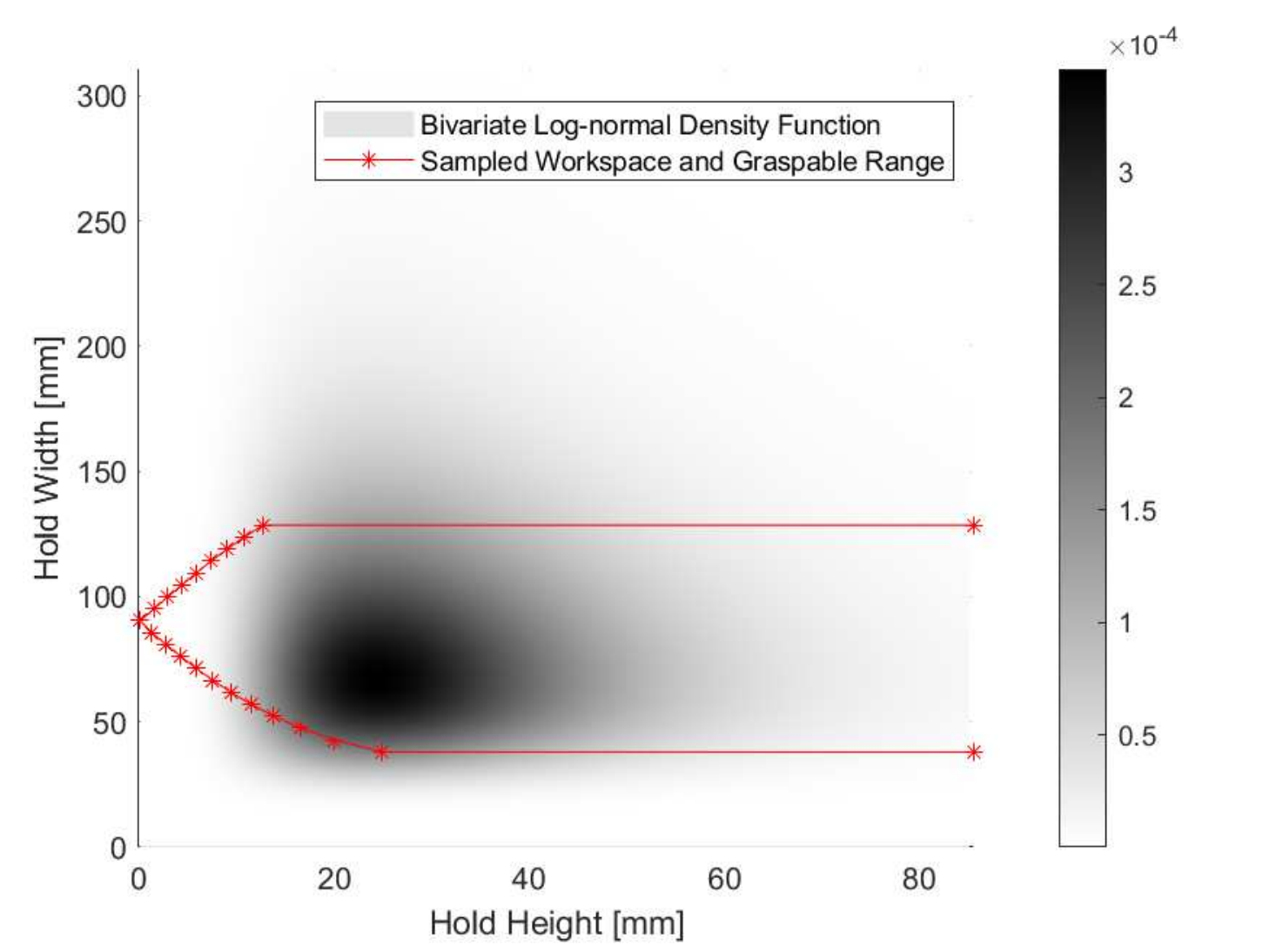}
    \label{fig:histogram}
    \caption{Bivariate log-normal distribution of bouldering hold sizes in the task environment. Gray scale represents the density and red points and lines are the boundary of the graspable object range. Any holds within the red boundaries are graspable under our assumptions as discussed in Section \ref{sec:workspace_samplling}. $\Omega_{l}$ and $\Omega_{u}$ determine the lower and upper bound of the graspable object width. The lower bound of the graspable hold height $H_i$ is computed by IK given uniformly sampled $\Omega_i$ between $[\Omega_{l}, \Omega_{u}]$. Our MOP objective function has successfully found the appropriate range that can cover $83.2$ \% of the black high density region while improving grasping forces and satisfying kinematic constraints. \label{fig:bivariate}}
    \label{leg1}
     \end{subfigure}
     \caption{The modeled task environment using the minimum bounding box and log-normal density function.}
\end{figure}

Since the environment is known, we define the task for GOAT as grasping bouldering holds.
We represent the environment as a bivariate Probability Density Function (PDF) of hold sizes. 
We consider the size of the minimum bounding box of each hold, width, length, and height.
Using a minimum bounding box simplifies the definitions of each object's shape and allows the robot to use a vision system to determine this minimum bounding box during operation to estimate gripping forces.
Since GOAT is a two-finger design, we redefine the lengths as width, representing the case when the gripper is oriented in 90 degrees. Hence, our PDF consists of the minimum bounding box 
with height on the $x$-axis and width on the $y$-axis. 
The density of the PDF represents the likelihood of each hold size.
We evaluate candidate distributions using a quantile-quantile plot. A log-normal (LN) best describes our skewed right density estimation histograms, shown in \fig{fig:histogram_height} and \fig{fig:histogram_width}.

A classical LN distribution is formulated in \cite{log_normal_57}. A bivariate LN distribution is denoted as \eq{eq:bvln}
is normally distributed in log domain, ${X}=\ln {Y}$, where $X$ is a set of positive, log-normally distributed variables with mean ${\mu_X}$ and standard deviation ${\sigma_X}$. In contrast, $Y$ is a set of positive, normally distributed variables with mean ${\mu_Y}$ and standard deviation ${\sigma_Y}$, and $\rho_{{Y}}$ is a single  correlation coefficient \cite{log_normal_history}. The analytical PDF solutions of the bivariate LN distribution is given in \cite{log_normal_57}. There is no closed-form solution of the Cumulative Distribution Function (CDF) for a correlated bivariate LN distribution. Generally, multivariate distributions do not have analytical equations of CDF, and thus we have to estimate the CDF of our distribution. 

In other environments (e.g., natural rock climbing), grasping object shapes can be observed using satellite or 3D scans. Our proposed objective function can work with any arbitrary PDFs with CDF estimations as long as they are in a continuous domain since we solve the optimization problem using Non-Linear Programming (NLP). 

\begin{figure}
    \centering
    \includegraphics[width=0.45\textwidth, trim = {0cm 0 0cm 0}, clip]{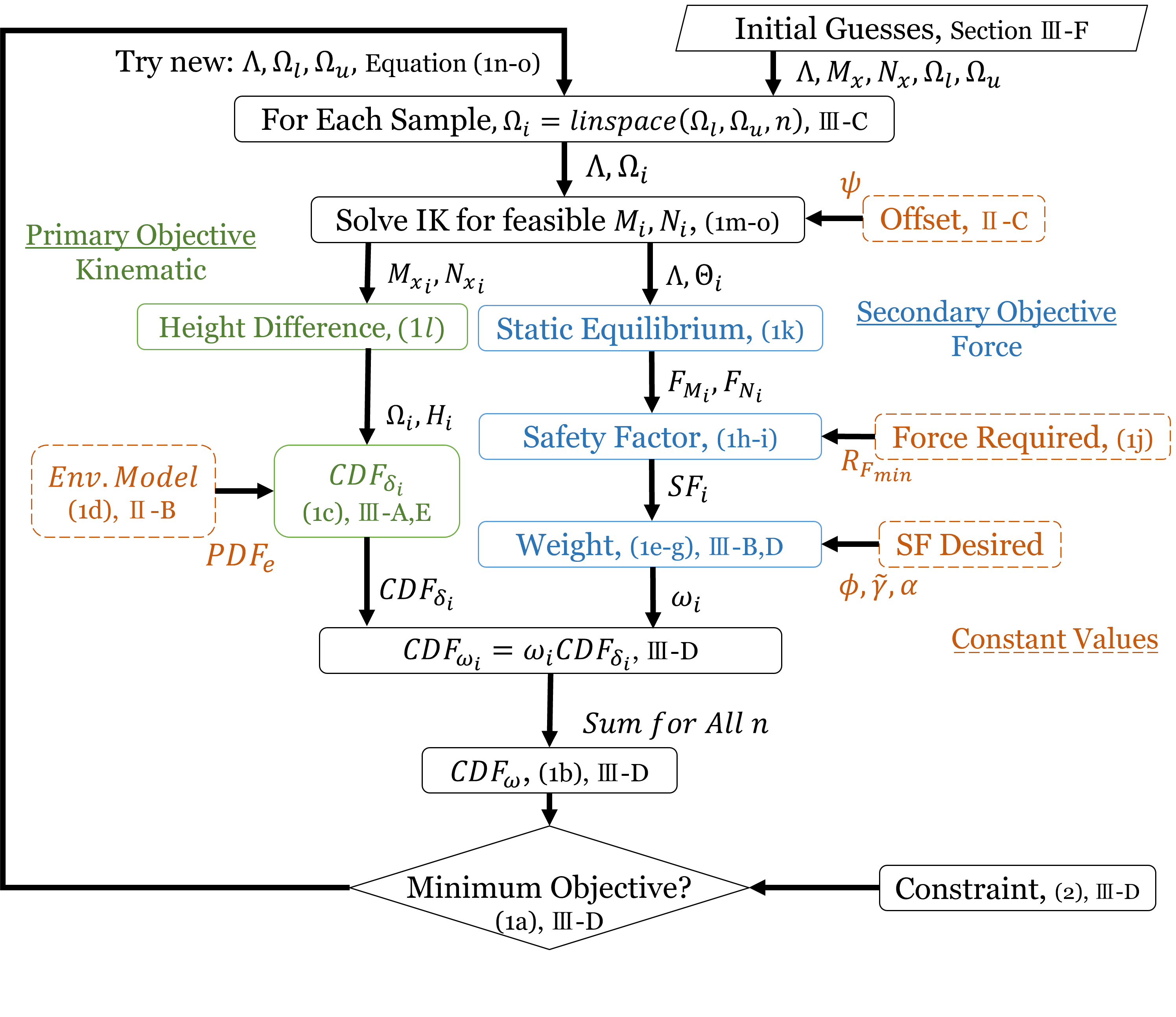}
    \caption{Our optimization algorithm flowchart. Each block describes a step and includes the corresponding equations and sections. We sample $\Omega_i$ that are linearly spaced between $\Omega_l$ and $\Omega_u$, the lower and upper bounds of the graspable object width. 
    We solve the inverse kinematics for feasible $M_i$ and $N_i$ contact points. Using IK solutions and given force requirements we compute a transmission ratio and a safety factor via static equilibrium analysis. The weighting $\omega_i$ is determined by a polynomial function that represents the relationship between a safety factor and adaptability. $CDF_\delta$ is a estimated CDF between each sample point $[\Omega_i,\Omega_{i+1}]$.
    The optimizer decides the appropriate kinematic range using the CDF of the environment model, which is weighted by $\omega$.} 
    \label{fig:flowchart}
\end{figure}


\subsection{Toe Accuracy of Position-controlled Limbed Robots \label{sec:offset}}
Since our application is climbing, the gripper needs to compensate for toe position errors. 
We assume that the robot targets the center of the bouldering hold when grasping. However, the limbed robot toe position may differ from the ground truth due to accumulated errors of joint angle controls, hardware constructions, localization, object position estimations, etc. 
Those errors can be estimated as a probability distribution through robot hardware experiments. 
We conservatively take a constant offset value, $\psi$, with a confidence interval given a PDF. This maximum offset from the center of the target object can result in missing the object since one finger close to the object surface touches first, and the other may not come into contact. GOAT can compensate this error as shown in \fig{fig:cad} and \fig{fig:gripper_parameters}.

\section{Grasping Kinematic and Force Optimization Using a Modeled Environment \label{sec:Opti}}
We optimize GOAT's linkage lengths to have an appropriate kinematic range and Transmission Ratio (TR) under constraints using NLP. Our objective function has to determine the bouldering hold ranges that GOAT can grasp while ensuring sufficient TR. The adaptability metric of the gripper is defined as the object sizes CDF. We divide our MOP into the single primary kinematic objective weighted by the secondary force objectives in \eq{eq:objective}. Our optimization algorithm is in \fig{fig:flowchart}.

\begin{figure}
\begin{subequations}
\label{eq:objective_all}
\begin{align}
\underset{\Gamma}{\operatorname{minimize}}\  \color{black} (1-{CDF}_{\omega})\label{eq:objective}\\
 \color{black}\operatorname{s.t.\ Constraints\ in\ (\ref{eq:all_const})}\nonumber\\
CDF_{\omega} =  \sum_{i=1}^{n}\left(\omega_{i} \cdot CDF_{\delta_i}\right)\label{eq:CDF_weighted}
\\
CDF_{\delta} = \iint_{\delta_{xy}} PDF_e \,dx\,dy \label{eq:cdf_sampled}\\
PDF_e = LN\left(\mu_Y, \sigma_Y, \rho_Y\right)  && 
\operatorname{(Env. model)}\label{eq:bvln}\\
\omega_i=-\frac{\alpha}{\kappa_{m_i}}-[\alpha\kappa_{p_i}]^{2}+1\label{eq:weighting_function} && 
\color{black}\operatorname{(Weighting)}\\
\kappa_{m_i} = (SF_i - \phi) &&\color{black}\operatorname{(SF\ lower\ bound)} \label{eq:sf_phi}\\
\kappa_{p_i} = SF_i - \tilde{\gamma} && \color{black}\operatorname{(SF\ target)}\label{eq:sf_gamma}\\
SF_i=\frac{{R_{F_i}}}{ {R_F}_{min}}&& \color{black}\operatorname{(Safety\ factor)}\label{eq:sf}\\
{R_{F_i}} = \frac{\norm{F_{N_i}-F_{M_i}}}{F_{actuator}} && \color{black}\operatorname{(Trans.\ ratio)}\label{eq:ratio_gripper}
\\
{R_F}_{min} = \frac{m_r}{\lambda F_{actuator}} && \color{black}\operatorname{(Required\ ratio)}\label{eq:ratio_min}\\
F_{N_i},F_{M_i}=SE(\Lambda,\Theta_i)&& \color{black}\operatorname{(Static\ equ.)}\label{eq:static_equ}\\
H_i = \abs{M_{x_i}-N_{x_i}} \label{eq:H_i} && \color{black}\operatorname{(Height\ difference)}\\
\Theta_i = IK(\Lambda,M,N)\label{eq:IK} && \color{black}\operatorname{(Inv.\ kinematics)}\\
M_i=\left[M_{x_i}, \frac{ { \Omega_i }}{2}+\operatorname{ \psi}\right] \label{eq:IK_contact_M} && \color{black}\operatorname{(Contact\ points)}\\
N_i=\left[N_{x_i}, -\frac{ { \Omega_i }}{2}+ \operatorname{ \psi }\right] \label{eq:IK_contact_N} && \color{black}\operatorname{(Contact\ points)}\\
\text{\eq{eq:IK_contact_M},\eq{eq:IK_contact_N} for }i=1,\dots,n \nonumber\\
\Gamma = \{\Lambda,M_{x_i},N_{x_i},\Omega_{l},\Omega_{u}\} && \color{black}\operatorname{(Decision\ variables)} \label{eq:decision}
\end{align}
\end{subequations}
\end{figure}


We define a decision variable set in \eq{eq:decision}, $\Gamma$, where $\Lambda$\ contains fifteen symmetric gripper link lengths. $M, N$ in \eq{eq:IK_contact_M}  and \eq{eq:IK_contact_N} are the gripper tip positions for all sampled kinematics points. $\Omega$ is a graspable object width, given an offset $\psi$ from the object central axis. The maximum and minimum $\Omega$ are decision variables which define the range of the graspable object size.

\subsection{A Metric for GOAT Adaptability}
Here we define a metric for the adaptability of GOAT. We consider the range of object sizes that GOAT can grasp for a given $\psi$ by passively shifting fingertips. 
As defined in Section \ref{sec:environment}, object sizes are described by their bounding box width and height. A wider width range of grasping implies greater adaptability, and thus an objective function should improve the gripper finger range in the $y$-direction. However, merely improving such a range does not guarantee a \textit{meaningful} adaptability enhancement. 
Adapting to infrequent object sizes may not be beneficial. Rather a gripper should cover the range of the object sizes that frequently appear in a task environment. Hence, we utilize the PDF that models our object environment to determine a useful grasping range in our objective function. The coverage of graspable object sizes is measured as a probability distribution using a CDF of the environment PDF. An ideal gripper that can grasp all holds in an environment will have a CDF of 1 with infinite link lengths. Nonetheless, the coverage is bounded and limited by kinematic constraints and the secondary force objective in Section \ref{sec:weight}.

\subsection{A Safety Factor-Based Auto-Tuning Weighting Function \label{sec:weight}}
It is essential to consider both the kinematic adaptability and the grasping force capacity when designing an adaptable gripper. 
One of the standard methods of a MOP is a weighted sum approach, which is sensitive to weight factors representing each objective's relative importance.
Adaptability and TR may conflict since improving one may adversely affect the other. Therefore, weightings need to be determined based on expertise and experience \cite{multiobject_design}. Here, we introduce an auto-tuning weighting function similar to the Carrillo's utility function \cite{auto-tuning}. We reduce our MOP into a single objective optimization weighted by the secondary objective. There exists a fundamental issue across MOPs, unit incompatibility. In gripper designs, we may include both kinematic and force units, and thus the weighting has to adjust to unit differences as well. Hence, we employ a Safety Factor (SF) of grasping force or TR, which is unitless. The TR represents the force ratio between inputs and outputs similar to a gear ratio. Note that our kinematic adaptability is represented as a CDF, which is unitless.

We set our primary objective to be the CDF of the graspable object range and the secondary to be a SF of TR because improving the CDF is always better and monotonic, whereas a higher SF only increases the margin of safety.
The secondary objective changes weighting of the corresponding CDF measurements based on a SF of the TR given the inverse kinematics (IK) solution since the relationship between the workspace and grasping forces are not constant. Therefore, those relationships can be expressed as follows:
\begin{itemize}
    \item Where Safety Factor is close to a lower bound of the SF, $\phi$, prioritize improving TR over adaptability. $\lim_{SF\to\phi} \omega(SF) = -\infty$
    \item Where Safety Factor is at a desired SF, $\gamma$, then prioritize improving adaptability over TR. $\omega(\gamma) = 1$
    \item Where Safety Factor is adequately higher than $\gamma$, then sacrifice TR to improve adaptability. $\lim_{SF\to\infty} \omega(SF) = -\infty$
\end{itemize}

Any decision variable sets, $\Gamma$, that result in a SF of less than the SF's lower bound should not be selected as GOAT's design, whereas $\Gamma$ with high SFs may not benefit our task but add more margin of safety.
TR will increase as the kinematics approaches to a singularity; hence, the higher SF is not preferred. 
The SF is computed by \eq{eq:sf} using TR required and TR based on tip reaction forces, \eq{eq:ratio_gripper} and \eq{eq:ratio_min}, respectively, where $\lambda$ is the Coulomb friction coefficient, $m_r$ is the robot mass. 



 \begin{figure}
    \centering
\includegraphics[width=.45\textwidth, trim = {1cm 0.05cm 0 .3cm}, clip]{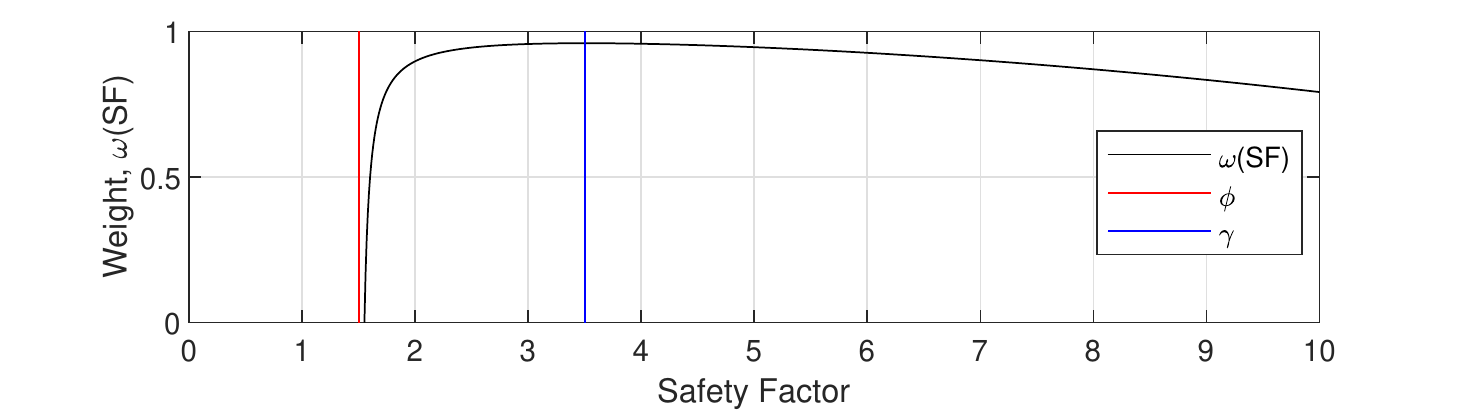}
     \caption{The plot of the factor-based weighting function, $\omega(SF)$ in \eq{eq:weighting_function}. Weight vs. a safety factor. The weight approaches to $-\infty$ at $SF=\phi$ and $SF=\infty$ and peaks around $\lambda$. \label{fig:weight}}
\end{figure} 

Such a nonlinear relationship can be approximated by a sum of two polynomial functions in \eq{eq:weighting_function}, where $\alpha$ controls the gradient of the weighting, $\kappa_{m_i}$ and $\kappa_{p_i}$ are in \eq{eq:sf_phi} and \eq{eq:sf_gamma}. The peak of this weighting function $SF(\gamma)$ is controlled by $\alpha$ and $\tilde{\gamma}$. The lower bound and the target sf $\phi$ and $\gamma$,respectively, are determined based on SF design criteria. 
Those constants specify the nonlinear relationship between kinematics and grasping force. 
This weight is strictly less than 1. The relationship between a SF and weighting is shown in \fig{fig:weight}.
We calculate a SF for each sampled IK solution using static equilibrium of GOAT in \eq{eq:static_equ}. 
The computed weights are applied to the corresponding $CDF_{\delta_i}$ in \eq{eq:CDF_weighted}, where $n$ is the number of the workspace sample points discussed in Section \ref{sec:workspace_samplling}.


\subsection{Workspace Sampling\label{sec:workspace_samplling}}
We optimize using the object environment PDF to achieve a meaningful grasping range.
$\Omega_{l}$ and $\Omega_{u}$ are decision variables representing the width of the narrowest and widest graspable object for a given $\psi$. We assume that a hold is graspable if its width is in the range and its height is taller than the contact point difference between $M_x$ and $N_x$. If an object height is shorter than the $H$ shown in \fig{fig:gripper_parameters}, then one of the fingers should miss the bouldering hold. This creates a lower boundary of graspable height. Any holds that are taller than $H$ is graspable as long as the robot arm can reach because we are only shifting the gripper in the negative $x$-direction. Hence, we set a constant $H_u$ to be the upper bound of graspable object.

We sample linearly spaced widths $\Omega_i$ between $[\Omega_{l},\Omega_{u}]$ for $i=1,\dots,n$, where n is the number of the samples. We solve for $M_{x_i}$ and $N_{x_i}$ using IK and contact point definitions in \eq{eq:IK}, \eq{eq:IK_contact_M} and \eq{eq:IK_contact_N}, where $\Theta$ is a set of joint angles. The differences between $M_{x_i}$ and $N_{x_i}$ represent the shortest hold graspable given $\Omega_i$ and $\psi$ as shown in \eq{eq:H_i}. Consequently, the set of $\Omega_i$ and $H_i$ represents the boundary of graspable range in \fig{fig:bivariate}. 


\subsection{Obtain $CDF_\omega$ from $CDF_{\delta_i}$ and $\omega_i$}
We compute $CDF_{\delta_i}$ in \eq{eq:cdf_sampled}, which is a CDF bounded rectangularly by [$\Omega_i$,$\Omega_{i+1}]$ and $[H_i,H_u]$. 
There is one exact IK solution given $\Omega_i$ and $\psi$ as described in Section \ref{sec:design}. Thus $H_i$, the shortest hold graspable, depends on the gripper design $\Lambda$.
The weighting $\omega_i$ in \eq{eq:weighting_function} is calculated for all $i$ using SF given the corresponding IK and static equilibrium solutions. In \eq{eq:CDF_weighted}, we discount respective $CDF_{\delta_i}$ by the force based weighting. Therefore, one design $\Lambda$ with a high $CDF_\omega$, meaning a large graspable object range, is penalized if the force SF is off the target SF. Maximizing the sum of all the weighted $CDF_{\delta_i}$ represents maximizing the graspable object range discounted by SF. The gripper that has the largest graspable object range and meets SF requirements at all sampled points will be the most optimal GOAT. Our objective function can find an appropriate $\Omega_l$ and $\Omega_u$ range even if their initial guesses are off from the high density region in \fig{fig:bivariate}.
We transform our maximization problem to minimization as shown in \eq{eq:objective} since the CDF and the weights are less than or equal to 1.

\subsection{Kinematic Constraints}
We have to grant kinematic constraints to satisfy loop closure and hardware limits. In this section, we cover each parameter definition and mathematical formulation that constraints \eq{eq:objective_all}. 

\begin{figure}[h!]
\begin{subequations}
\label{eq:all_const}
\begin{flalign}
    \norm{D_{u}-D_{l}} < \epsilon_{{th}} && \text{(Kinematic loop)}\label{eq:kin_loop}\\
    M_x, N_x < Tip_{u} && \text{(Tip contact points)}\label{eq:const_tip}\\
    \norm{D_{max}-D_{min}} \leq Stroke_{ac} && \text{(Actuator stroke)}\label{eq:const_stroke}\\
    L_{l} < L_j < L_{u} \text{ for } j = 2,\dots,15 &&  \text{(Topology)}\label{eq:const_topo}\\
    0 < L_1 < L_{u} && \text{(Link 1)} \label{eq:const_link1}\\
    \theta_l < \theta_3, \theta_6 < \theta_u &&  \text{(Joint angles)}\label{eq:const_theta3}
\end{flalign}
\end{subequations}
\end{figure}

\subsubsection{Kinematic Loop Closure Constraints}
Our whippletree gripper design contains a closed-loop structure, described by the connecting links, $L_{2,\dots,7}$, defined in \fig{fig:gripper_parameters}. 
We define the closed-loop as an equivalent tree structure that is acquired by cutting the closed-loop structure at the point D. The equivalent tree structure consists of two branching link paths defined by the $L_{2,\dots,4}$ and the $L_{5,\dots,7}$ serial chain. The geometric constraint in \eq{eq:kin_loop} is required to incorporate the loop-closure constraint into the kinematic solution.



\subsubsection{Tip Contact Points Constraints}
Tip contact points are computed by solving the IK for given contact surface functions for the point M and N. The distance to the tip contact points from the end of the robot toe position is bounded by the robot shoulder servomotor continuous torque since this distance is virtually extending the robot leg length. Thus, \eq{eq:const_tip} should be satisfied, where $Tip_{u}$ is the upper bound of the tip distance allowed.

\subsubsection{Actuation Stroke Constraints}
As shown in Section \ref{sec:design}, an input force to GOAT is applied at the point D in the negative $x$-direction. We employ a DC linear actuator as our input since it can provide a significantly higher force at stall than rotational actuators due to its high gear ratio. However, a DC linear actuator is limited in stroke, or the actuator length becomes consequentially long. 
This creates a constraint on the point D motion range. In \eq{eq:const_stroke}, $D_{max}$ and $D_{min}$ are the point D positions, $[x,y]$ when fully opened and closed, respectively.

\subsubsection{Topology Constraints}
Since the gripper topology is defined in our design optimization as a whippletree-based linkage mechanism, all link length, except $L_1$ cannot be zero or near zero. 
Therefore we impose the lower and upper bound of the link length, $L_l$ and $L_u$ in \eq{eq:const_topo} to maintain the design and limit to a finite length. The topology of the whippletree and the linkage system does not change with $L_1 = 0$, and thus \eq{eq:const_link1}. Two joint angles in \eq{eq:const_theta3} needs to be constrained by the upper and lower bound of $\theta_u$ and $\theta_l$ so that $L_3$ and $L_6$ point toward the positive $x$-direction to satisfy the requirement of a whippletree mechanism. 


\subsection{CDF Estimation via Riemann Summation \label{sec:cdf_delta}}
A closed-form CDF is not known in general, but we need to estimate $CDF_{\delta_i}$ in \eq{eq:cdf_sampled}. 
We use Riemann summation to estimate the CDF of the gripper adaptability coverage deterministically for computational simplicity. We compute uniform sample point sets between $\{[H_i,H_{u}], [\Omega_i,\Omega_{i+1}]\}$, which represents the graspable bouldering holds discussed in Section \ref{sec:workspace_samplling}. Riemann summation is sufficient for our one modal LN distribution and more accurate CDF estimations do not alter the link lengths significantly. 


\subsection{Nonlinear Programming Initial Guess Problems \label{sec:initial_guess}}
NLP solvers begin searching for a minimum from initial guesses of decision variables, but are not guaranteed to reach a global minimum or a feasible solution given a feasible problem. One approach to this problem is starting from different guesses for each optimization run, then adapt a solution with the minimum objective value among the solution sets. In design optimizations, random link length guesses are unlikely to be a valid design, thus we apply different valid link length sets. 
This helps the solvability problem in Section \ref{sec:workspace_samplling}.


\section{RESULTS}\label{results}
\begin{figure}
    \begin{subfigure}{0.24\textwidth}
    \centering
\includegraphics[width=\textwidth,trim = {0.5cm 0 1cm 0}, clip]{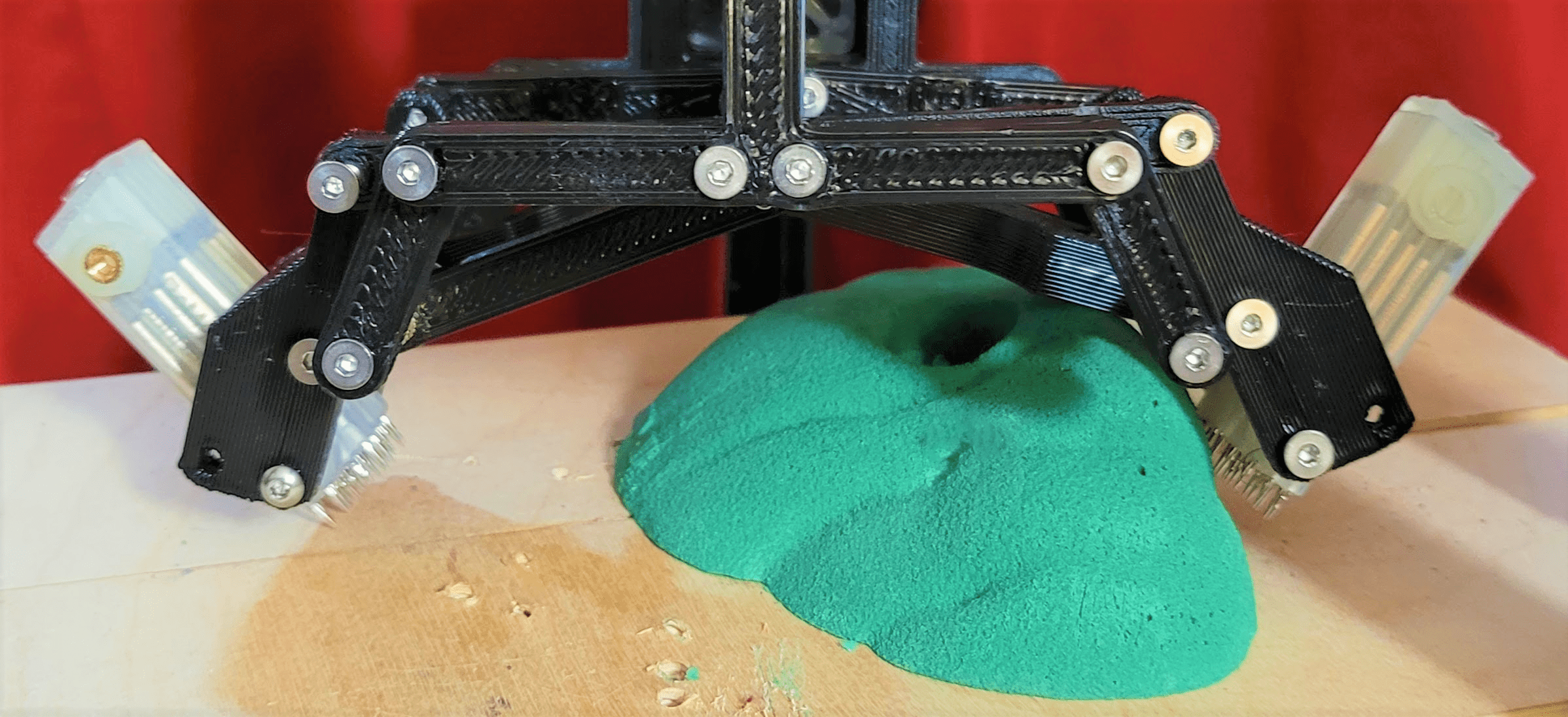}
    \end{subfigure}
     \begin{subfigure}{0.24\textwidth}
         \centering
         \includegraphics[width=\textwidth,trim = {0.5cm 3cm 1cm 1cm},clip]{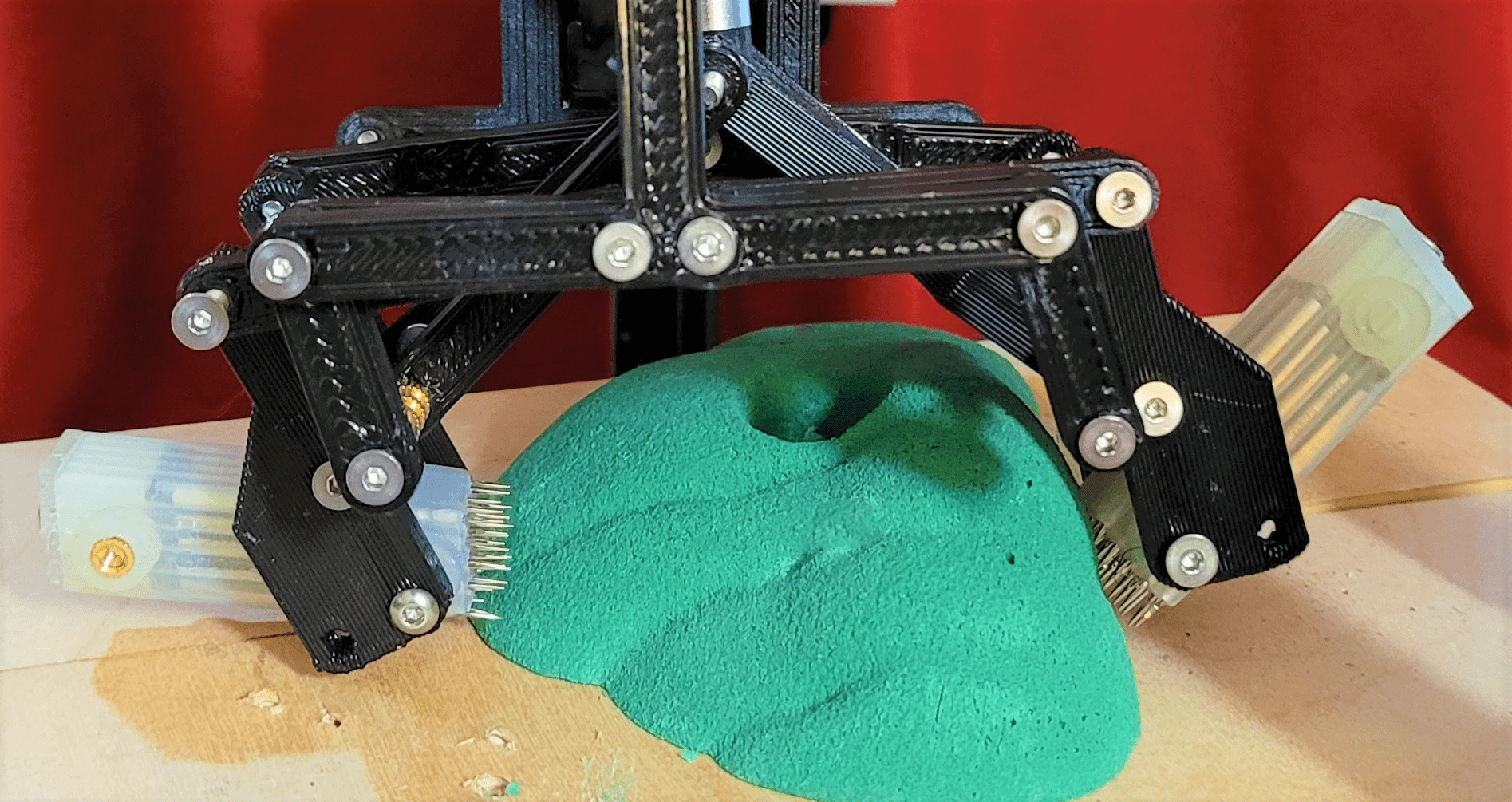}
     \end{subfigure}
     \caption{An example of grasping an irregular and off-center object. The right fingertip touches the bouldering hold first in the left figure, but GOAT balances the finger load passively and successfully grasps the object as shown on the right.\label{fig:opti_config}}
\end{figure}

This section presents implementation and verification of our objective function for GOAT. 
Constraints and parameters are determined. 2D models of our optimal gripper are shown in \fig{fig:cad} and one example of off-center bouldering hold grasping is shown in \fig{fig:opti_config}. The robot weight creates a moment arm at GOAT in 2D space depending on the robot body distance from the wall. Since the GOAT linear actuator is not back driable the pulling force is critical than the force in a gravity direction.
We evaluate the maximum pulling force required to detach our 2D and 3D gripper designs to determine slip between a bouldering hold surface and spines. 

\subsection{Kinematic Constraints and Variables}
Kinematic constraints and variables are in Table \ref{tb:opti_params}.
The offset $\psi$ is based on the toe position error of $4$ mm on one leg from our two-wall climbing hexa limbed robot, SiLVIA \cite{silvia}. However, this analysis does not include observation noise, localization errors, or effects from other legs. Consequently, we set three times more conservative value in our optimization. 
 The $L_{u}$ and $L_{l}$ are defined to fit in our 3D printer and to leave room for two-pin joints.

\subsection{Optimized Gripper Design and Theoretical Adaptability}
Our objective function is evaluated by Ipopt solver \cite{ipopt} with CasAdi \cite{casadi}. The optimized link length set $\Lambda$ is listed in Table \ref{tb:opti_link}. The optimal design has the theoretical graspable range $CDF = 83.2 \%$. The optimal minimum and maximum width range is found to be $\Omega [\operatorname{min,\ max}] = [38.02\text{ mm}, 128.5\text{ mm}]$. Each configuration is shown in \fig{fig:opti_config}. The theoretical graspable bouldering hold range is plotted in \fig{fig:bivariate}. The optimal design covers the majority of the dense regions in the PDF. 

Ipopt runs a total of 300 times with different initial guesses of three linkage length sets. Each linkage length initial guess set is linearly amplified from $0.05$ to $2.00$ by ten along with pre-calculated IK solutions, which is also amplified accordingly. 
In our implementation, the optimizer returned five identical solutions with a minimum objective out of nine solutions.
The gripper workspace is sampled with $n = 20$ using IK and the object model PDF is sampled at $4000$ points. More sample points can improve accuracy of the CDF estimation, but optimal parameters do not change significantly above a sufficient amount of samples.

\begin{table}[]
\sisetup{table-format=1.4e-1,exponent-product = \cdot}
\centering
\caption{Optimization parameters.\label{tb:opti_params}}
\begin{tabular}{llllll}

$\psi$ & $12.0$ mm & $\Phi$ & $1.5$ & $\gamma$ & $3.5$ \\ \hline
$Tip_u$ & $50$ mm & $Stroke_{ac}$ & $30$ mm & $L_u,L_l$ & $10, 200$ mm\\ \hline

\end{tabular}
\end{table}

\begin{table}[]
\centering
\caption{Optimal link lengths in mm.\label{tb:opti_link}}
\begin{tabular}{llllll}
\hline

$L_1$         & $0.00$  & $L_2,L_7$  & $37.52$     & $L_3,L_6$  & $24.52$  \\ \hline
$L_4,L_5$      & $50.07$  & $L_8,L_{11}$    & $43.26$  & $L_9,L_{12}$ & $24.26$  \\ \hline
$L_{10},L_{13}$    & $10.00$  & $L_{14},L_{15}$   & $17.20$  &        &  \\ \hline
\end{tabular}
\end{table}



\subsection{Spine Fingertips}
A spine gripper has been successful for climbing purposes where minor distractions are acceptable \cite{climb_yuki}. 
A spine cell based on \cite{spine_cell} is designed with twenty-five $\diameter 0.93$ spines, each loaded by a $5$ mN/mm spring in one cell at $7.5 \degree$. One cell is rigidly attached at each tip of the gripper as a part of $L_{14}$ and $L_{15}$ at a sufficient angle to avoid collisions with the ground. 
In our experiments, we are interested in measuring the maximum force the gripper can withstand until it detaches from a fixed rock.
However, a sharp spine can penetrate deeply into the a polyurethane-mixture bouldering holds. Then the gripper does not detach unless either the polymer hold surface structure or the spines are destroyed or bent, which requires a substantial pulling force. We rounded the spine tip by scratching them on sandpaper. This effectively prevents spines from penetration, but they can still insert into the micro cavities on the hold surface, which causes friction \cite{climb_yuki}.
 The aged spines effectively increase the friction coefficient between the fingertips and the hold, and the results obtained by this method represent the gripper performance rather than material strength limits.

\subsection{Gripping Force Test Bed}
\begin{figure}
    \begin{subfigure}{0.2\textwidth}
    \centering
\includegraphics[width=\textwidth, trim = {0cm 0 0cm 0.1cm}, clip]{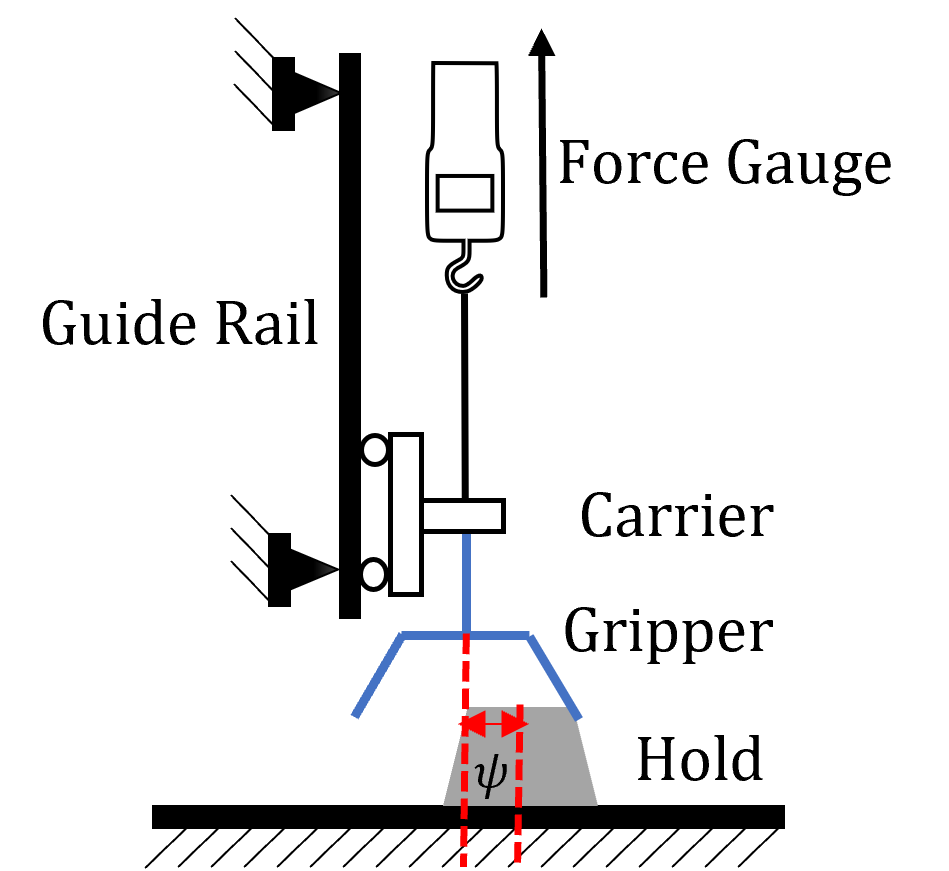}
\caption{Experiment setup for the maximum pulling force measurements. The force gauge is pulled in the arrow direction and the peak force is recorded. Each bouldering hold is installed with $\psi$.\label{fig:testbed}}
    \end{subfigure}
     \begin{subfigure}{0.23\textwidth}
         \centering
   \includegraphics[width=0.5\textwidth,trim = {1cm 3cm 0cm 1cm}, clip]{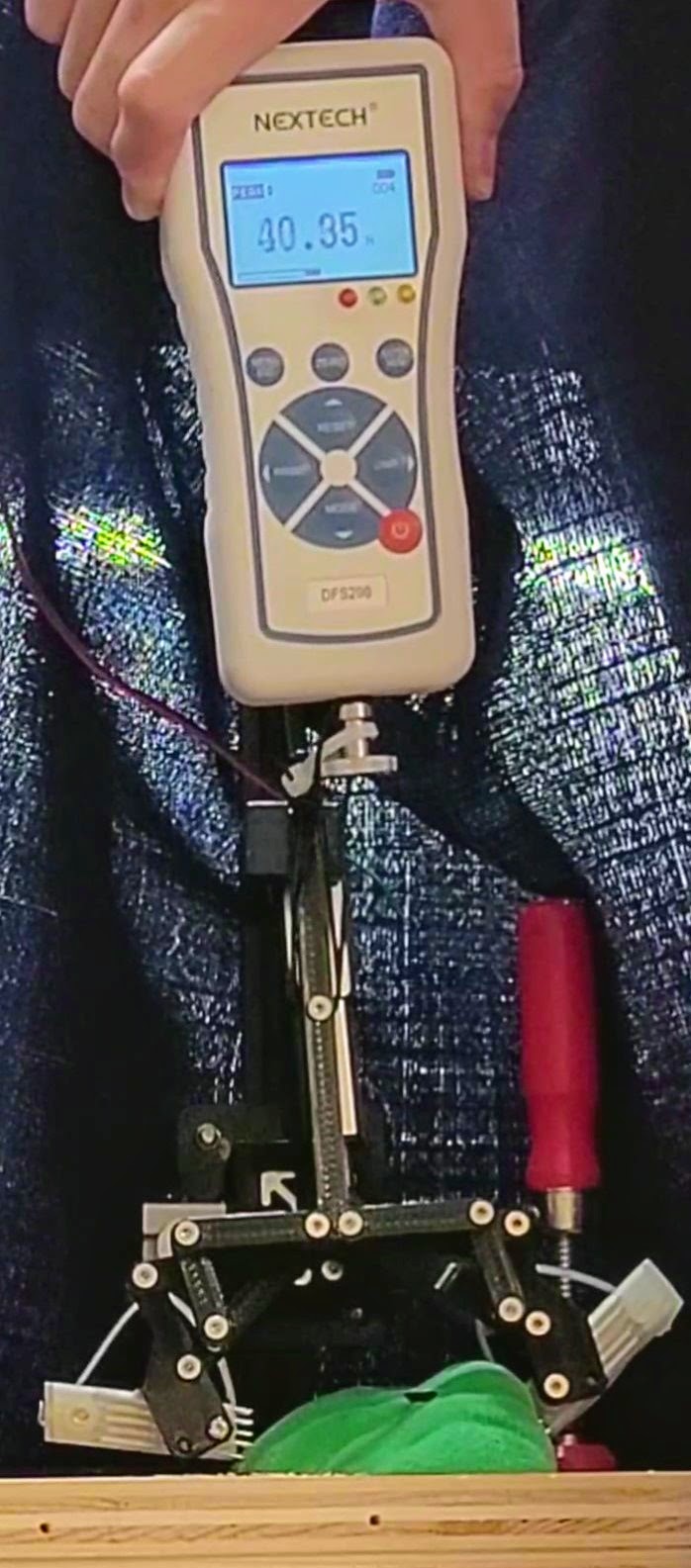}
   \caption{Our under-actuated GOAT gripper in the test bed.}
     \end{subfigure}
     \caption{Test bed configuration and our 3D printed GOAT.}
\end{figure}

Our grasping force testing setup is shown in \fig{fig:testbed}. The gripper is fixed to a linear rail carrier off-center by $\psi$, and it can only move in the direction indicated by the arrow. A bouldering hold is secured onto the test bed so that the gripper fingertips approach the points corresponding to its measured bounding box width. Each hold is tested twenty times for both the width and the length. The gripper linear actuator is supplied by an external DC power source and the maximum pulling force is recorded with a force gauge. Outliers are removed from the data.



\subsection{Adaptability and Grasping Force Evaluations \label{sec:evaluation}}
\begin{figure}
    \centering
    \includegraphics[width=0.4\textwidth,trim = {0.5cm 0 0.5cm 0}, clip]{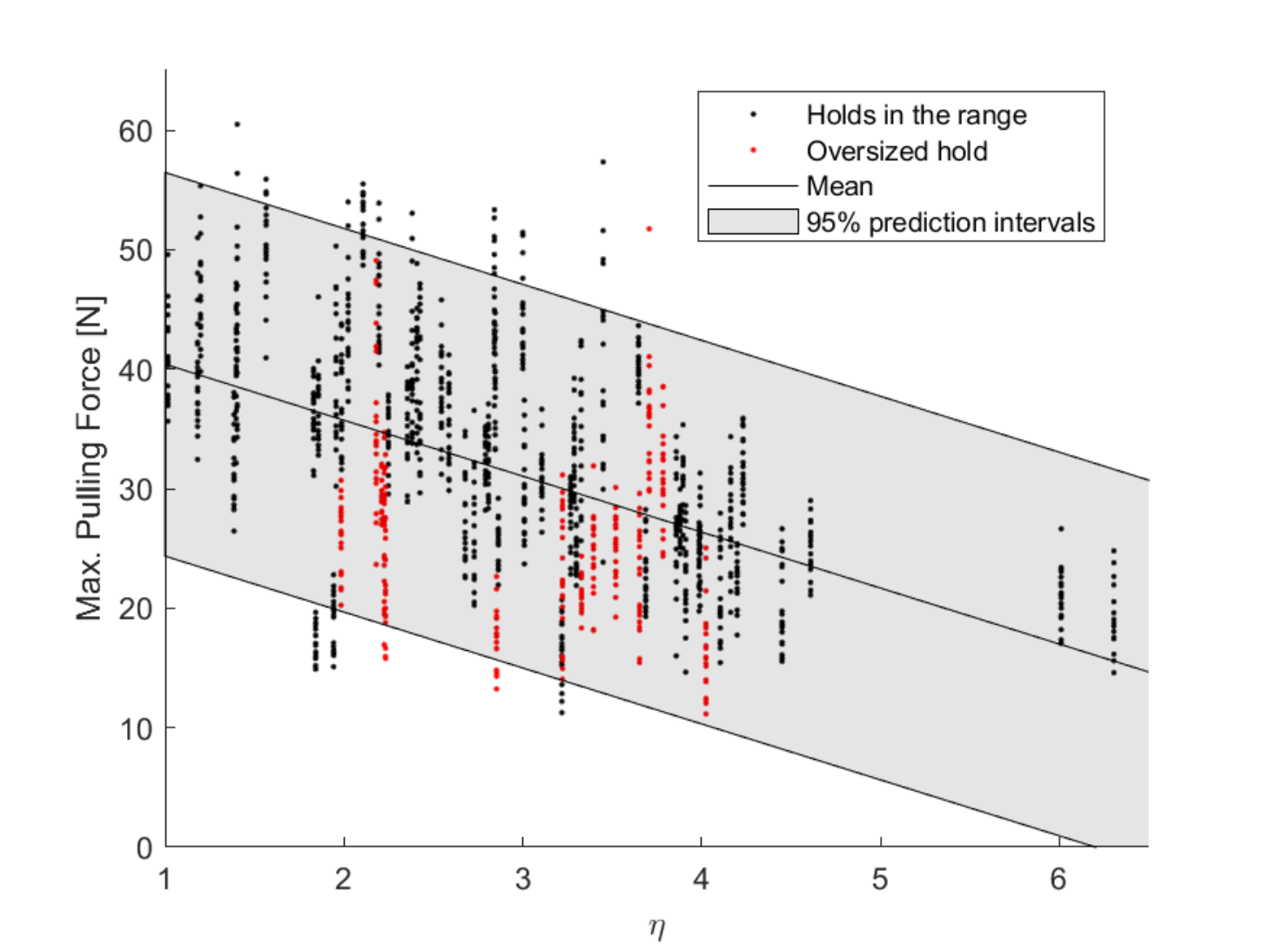}
     \caption{The optimal gripper maximum pulling force analysis. 48 unique bouldering holds are tested in two orientations, width and length directions, twenty times each. Width-Height ratio, $\eta$, represents the slope of the bouldering hold surface. As the ratio gets closer to 1, the maximum pulling force increases. GP is performed and the mean forces and $95$  \% confidence interval are determined. The gripper needs to withstand at least $13.3$ N}\label{fig:result_modeled}
\end{figure}

 The pulling forces for a total of 96 unique bouldering hold sizes with aging processed spines are shown in \fig{fig:result_modeled}. 
 
 %

\begin{figure}
\begin{subfigure}{0.24\textwidth}
         \centering
   \includegraphics[width=\textwidth,trim = {2cm 0cm 2cm 0.5cm}, clip]{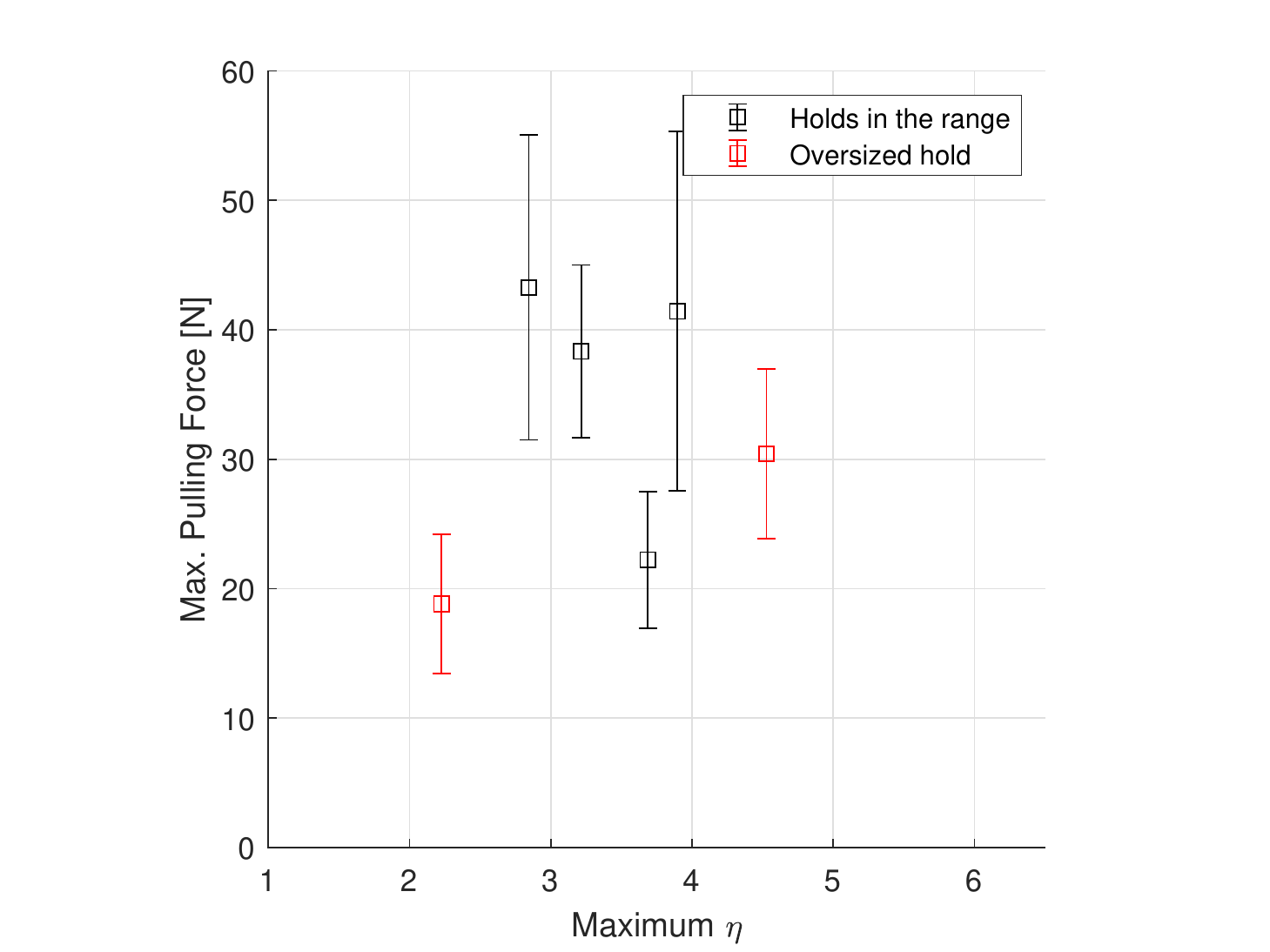}
   \caption{3D design test results against the maximum $\eta$.\label{fig:result_3d_eta}}
     \end{subfigure}
    \begin{subfigure}{0.24\textwidth}
    \centering
\includegraphics[width=\textwidth, trim = {2cm 0 2cm -0.25cm}, clip]{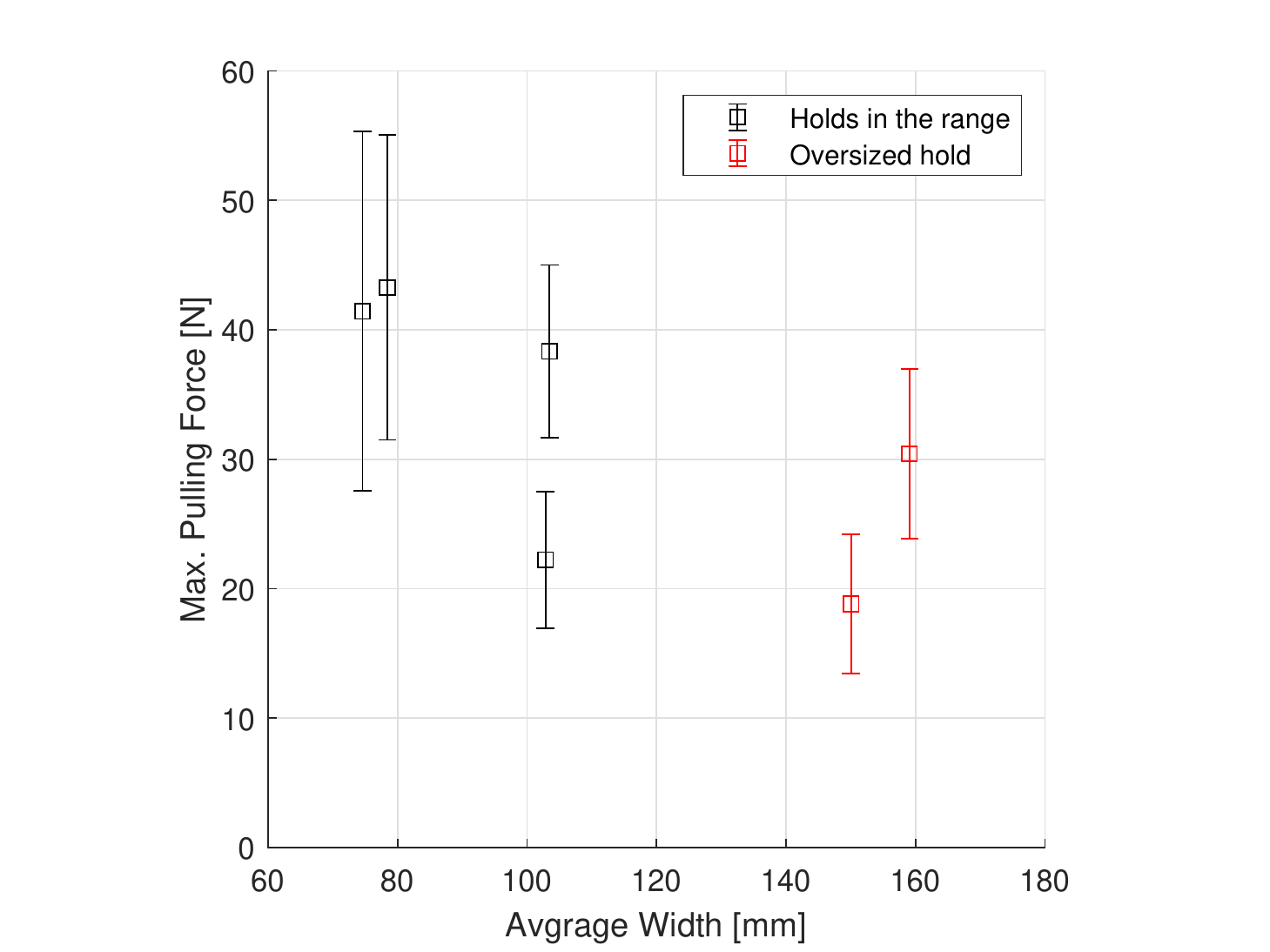}
\caption{3D design test results against the average of the minimum bounding box width and length.\label{fig:result_3d_avg}}
    \end{subfigure}
     
     \caption{3D design max pulling force evaluation. Six different size holds are tested and we take the average of the minimum bounding box width and length and the maximum $\eta$. Two large holds in red are outside of the graspable range.\label{fig:3D} }
\end{figure}

Our environment is modeled based on the minimum bounding box as defined in Section \ref{sec:environment}. This assumption is conservative since bouldering holds are similar to concave shapes. Therefore GOAT can potentially grasp oversized holds. 
Assuming that the holds are semi-ellipsoids, we have observed that the maximum pulling forces increase as the ratio of the width and height, $\eta$ goes to 1. In most cases, grasping forces of oversized holds were lower than means of grasping forces at corresponding $\eta$. 
The gripper cannot enclose the hold within the fingertips, and oversized holds have shallower slope at contact points between semi-ellipsoid surface and fingertips than calculated $\eta$.
Hence, spines are inserted in the surface cavity at a shallow angle against the pulling force. Nonetheless, it is beneficial to evaluate out-of-range hold grasping forces for a case where a wall climbing robot is allowed to take some risks in planning \cite{risk_aware}. A chance-constrained planner may decide to grasp an oversized hold when relatively high failure probability is permitted to enhance the graspable hold range. Any holds narrower than the graspable range are not evaluated. 

In this experiment, $79.2 \%$ of holds are graspable by adapting to unique bouldering hold geometries and compensating the offset $\Phi$.
This is less than the theoretical bound since some holds are close to a crescent shape, resulting in an actual shortest width significantly different from the bounding box assumption. 
The fingertips get caught in holds with concave surfaces for human grasp positions, and do not detach unless structures are physically damaged. Hence, such bouldering holds are not included in the results but counted as graspable. 

\subsection{Stochastic Grasping Modeling via Gaussian Process \label{sec:GP}}
Grasping forces of the spine grippers on rocky surfaces are stochastic since each spine inserts in tiny cavities but how well they are inserted is not constant over every trial \cite{climb_yuki}. Such stochastic forces can be modeled via Gaussian Process (GP) as demonstrated in \cite{risk_aware}. 
GP is performed with a linear kernel to output the maximum pulling forces given $\eta$. Zero force should be expected if the width is less than $\Omega_{l}$, which is a constraint. The gripper should withstand at least $13.3$ N for our climbing robot. Though the mean pull forces are above this limit for the range of our tests, at $\eta \approx 3.35$, the lower bound of the $95$ \% confidence interval reaches this limit. Consequently, the robot has to avoid $\eta > 3.35$, or the planner potentially needs to take some risks. The robot vision system can use the minimum bounding box to estimate the maximum pulling forces with this GP model. GOAT can compensate the toe position uncertainty and the diversity of the bouldering hold geometries.

\subsection{3D Design Gripper Grasping on the Robot \label{sec:3D}}
We can combine two instances of GOAT perpendicular to each other to form a 3D design. For 3D design, the environment definition is still the minimum bounding box, but now including length. However, in Section \ref{sec:environment}, we have accounted length as width in our object environment model. Therefore our optimized 2D design is optimal in both width and length direction and all the other conditions such as toe position accuracy assumption, force requirements, and kinematic constraints remain the same.
Hence we can apply the same optimal parameter, $\Gamma$ for both GOATs.
They are stacked on top of each other and link shapes are modified to avoid collisions while maintaining the same joint positions. One set of the fingertips are higher than the other since our climbing robot only consists of 4 DoF and the gripper may approach to a hold at angle. This design prevents spine cells from hitting the wall first.

The 3D version is tested with the same testbed and several bouldering holds are selected: two relatively small, around the mean of the range, and oversized to verify that our 3D versions can grasp holds as well as 2D design.
This 3D design helps stability, likelihood of successful grasping, and improves graspable hold ranges since the 3D version can grip a bouldering hold if one of the side lengths is within the range. There is no notable increase in the maximum pulling force because adding more two-finger grippers only increases contact points. \fig{fig:3D} shows that our 3D version design can withstand sufficient forces for climbing, but the maximum pulling force is lower for oversized holds and larger $\eta$ as discussed in Section \ref{sec:evaluation}.

\section{CONCLUSION and FUTURE WORK}\label{conclusion}
In this paper, we presented the rigid whippletree-based gripper GOAT and its MOP design optimization. Our proposed rigid gripper can adapt to an object and compensate position control errors.
Our proposed multi-objective function simultaneously improves the adaptability and force TR of the gripper with an auto-tuning weighting function. 
The proposed multi-objective function has successfully determined an optimal grasping range, which concurrently enhances the gripping force, by using the object environment's mathematical model.
Our bouldering hold grasping tests demonstrate that our rigid, durable gripper provides sufficient grasping forces for one-wall climbing tasks. 



Future work will aim to reduce the assumption of the minimum bounding box to include 3D geometries of each object in our object environment PDF. This will allow us to optimize other parameters, such as contact angle.






\end{document}